\pdfoutput=1
\documentclass{article}

\PassOptionsToPackage{numbers,sort&compress}{natbib}
\usepackage[main, final]{neurips_2026}

\usepackage[utf8]{inputenc} %
\usepackage[T1]{fontenc}    %
\usepackage{hyperref}       %
\usepackage{url}            %
\usepackage{booktabs}       %
\usepackage{amsfonts}       %
\usepackage{nicefrac}       %
\usepackage{microtype}      %
\usepackage{xcolor}         %
\usepackage{placeins}
\usepackage{flafter}

\title{Matching-Based Few-Shot Semantic Segmentation Models Are Interpretable by Design}

\author{%
  Pasquale De Marinis\thanks{Correspondence to \texttt{pasquale.demarinis@uniba.it}.} \\
  Department of Computer Science \\
  University of Bari Aldo Moro \\
  \AND
  Uzay Kaymak \\
  Jheronimus Academy of Data Science \\
  Eindhoven University of Technology \\
  \And
  Rogier Brussee \\
  Jheronimus Academy of Data Science \\
  Eindhoven University of Technology \\
  \AND
  Gennaro Vessio \\
  Department of Computer Science \\
  University of Bari Aldo Moro \\
  \And
  Giovanna Castellano \\
  Department of Computer Science \\
  University of Bari Aldo Moro \\
}

\usepackage{svg}
\usepackage{multirow}
\usepackage{booktabs}
\usepackage{colortbl}
\usepackage{url}
\usepackage{makecell}
\usepackage{tabularx}
\usepackage{doi}

\definecolor{lightgray}{gray}{0.9}
\definecolor{lightgreen}{rgb}{0.9,1,0.9} %

\newcommand{\forward}[1]{\cellcolor{lightgray}#1}
\usepackage{amsmath}
\usepackage[capitalise]{cleveref}
\hypersetup{pdftitle={Matching-Based Few-Shot Semantic Segmentation Models Are Interpretable by Design},
  pdfauthor={Pasquale De Marinis, Uzay Kaymak, Rogier Brussee, Gennaro Vessio, Giovanna Castellano}}
\definecolor{mydarkblue}{rgb}{0,0.08,0.45}  %
\hypersetup{colorlinks=true, linkcolor=mydarkblue, citecolor=mydarkblue, filecolor=mydarkblue, urlcolor=mydarkblue}

\begin{document}

\maketitle

\begin{abstract}
Few-Shot Semantic Segmentation (FSS) models achieve strong performance in segmenting novel classes with minimal labeled examples, yet their decision-making processes remain largely opaque. Explainable AI has matured on standard computer vision tasks, but not on FSS, where an explanation would tell a practitioner which support pixels the prediction rested on and which examples to provide next. We argue that matching-based FSS models are interpretable by design: their query--support similarity scores, once normalized and processed, attribute the prediction to individual support pixels. Our approach, Affinity Explainer (AffEx), makes this intrinsic interpretability explicit by extracting attribution maps directly from matching scores at multiple feature levels, without requiring gradients or external perturbations. We extend the standard causal metrics for evaluating explanations to the FSS domain and propose additional metrics to better capture the practical utility of explanations in few-shot scenarios. Experiments on two FSS benchmarks, covering seven architectures that range from classical matching-based models to recent pipelines built on vision foundation models (DINOv2/v3, SAM, SAM2) and a prototype-based head, show that AffEx outperforms adapted standard attribution methods at nearly every evaluation point, and that the gain comes from the matching mechanism itself. On pipelines whose segmentation stage transforms the matching output, the same evaluation doubles as a diagnostic of where the decision is made, and on the prototype-based head AffEx subsumes the prototype-style read-out as a special case. Qualitatively, the attribution maps expose how each architecture resolves the support set, and we show how they guide model diagnosis. Code is available at \url{https://github.com/pasqualedem/AffinityExplainer}.

\end{abstract}

\section{Introduction}
\label{sec:introduction}

\begin{figure*}[!t]
    \centering
    \includegraphics[width=\linewidth]{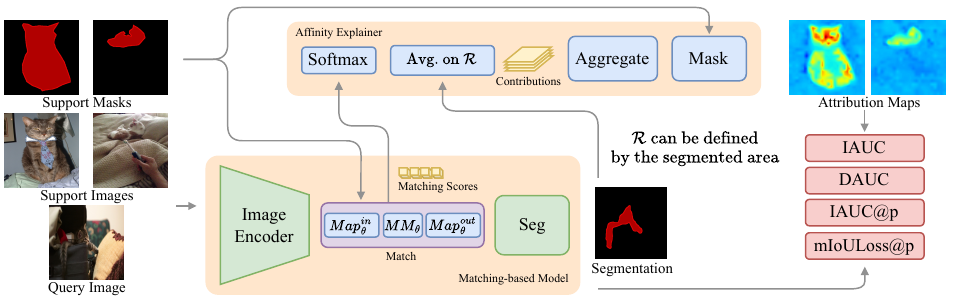}
    \caption{The proposed framework (two examples, one class). AffEx applies to any FSS model adhering to the illustrated structure: it reads the matching scores produced by the matching mechanism $\text{MM}_\theta$, before $\text{Map}^{\text{out}}_\theta$, normalizes the scores of each query position with a softmax over the support, averages the result over a region $\mathcal{R}$ (the predicted mask, the ground truth mask, or a user-specified region), and aggregates the per-level maps; the support masks assign the attribution sign. The resulting maps highlight the support regions most influential in segmenting the query.}
    \label{fig:framework}
\end{figure*}

Deep models have achieved remarkable performance in areas such as image recognition, speech processing, and language translation, yet their internal reasoning remains opaque, making them ``black boxes'' in domains requiring accountability, such as healthcare and law. Explainable AI (XAI) aims to address this issue, and significant progress has been made in conventional tasks such as image classification and language modeling. Nevertheless, explainability in Few-Shot Learning (FSL), which seeks to generalize from a handful of labeled examples per class \citep{wang2020generalizing}, remains underexplored, with most FSL studies focusing on accuracy rather than on explaining the models' predictions.

In \textit{Few-Shot Semantic Segmentation} (FSS), in particular, every prediction is conditioned on the examples: they provide the model with the only definition of the class to segment. When a segmentation fails, it is unclear whether the issue lies in the model, the choice of support images, or the task itself, and the question a practitioner needs answered is which pixels of the support set taught the model what it did. Existing analyses of few-shot segmentation models are query-centric, visualizing where the query image matches the support \citep{fan_self-support_2022, boudiaf_few-shot_2021, minHypercorrelationSqueezeFewShot}; to date, no dedicated work attributes the prediction back onto the support set itself. Nor does the standard XAI toolbox transfer to this question: as we show in \cref{sec:experiments}, most gradient- and perturbation-based attribution methods, adapted to the few-shot setting and evaluated causally, cannot surpass even a trivial baseline that returns a noisy copy of the support mask, and on modern pipelines built on vision foundation models (VFMs), gradient-based attribution is mostly not applicable, their decision paths containing discrete, non-differentiable steps.

We argue that matching-based FSS models are \emph{interpretable by design}: the signal needed to answer the question is already inside the models, in the matching scores that these architectures \citep{zhangFewShotSegmentationCycleConsistent2021, minHypercorrelationSqueezeFewShot, langLearningWhatNot2022, shiDenseCrossQueryandSupportAttention2022, zhangEfficientSamplingbasedGaussian2025} compute between support and query features. We introduce Affinity Explainer (AffEx), which \emph{reads} these scores from the model's forward pass, without modifying the model, and turns them into support-side attribution maps, highlighting the support pixels that most strongly influence the segmentation of the query, without gradients, external perturbations, or any training. The overall framework is illustrated in \cref{fig:framework}.

The contributions of this paper are as follows.
\begin{itemize}
    \item We formalize support-side attribution as a task: an attribution function that maps a region of the query prediction onto the pixels of the support set that determined it.
    \item We introduce AffEx, which solves this task by reading the matching scores the models already compute. It needs no training, no gradients and a single forward pass, and obtains the best faithfulness scores in nearly every setting we evaluated.
    \item We build the causal evaluation the task requires, extending classical insertion/deletion metrics to perturb the support set rather than the query, together with early-curve metrics that isolate the decision-relevant regime.
    \item We provide a benchmark over seven architectures (classical matching-based models, four recent pipelines built on vision foundation models, and a prototype-based head on which AffEx subsumes the prototype-style read-out as a special case), showing that post-hoc XAI largely fails or is inapplicable, and that the same evaluation doubles as a diagnostic of whether a model's decision is carried by its matching stage or by its segmentation stage.
\end{itemize}

\section{Related Work}
\label{sec:related_work}

\subsection{Few-Shot Semantic Segmentation}

FSS combines semantic segmentation, which assigns a class label to each pixel, with FSL, addressing scenarios where the large annotated datasets required by conventional segmentation models are impractical to obtain \citep{finn_model-agnostic_2017, snell_prototypical_2017, sung_learning_2018, vinyals_matching_2017}.

In an $N$-way $K$-shot ``episode'', the model receives a support set $\mathcal{S}$ containing $K$ annotated examples per class and a query image $I_q$ that may contain up to $N$ target classes, and must predict the corresponding segmentation mask; performance is averaged over episodes sampled from disjoint class sets to evaluate generalization to unseen categories.

Early FSS approaches relied on prototype learning, where prototypes are formed via masked average pooling and compared to query features \citep{wangPANetFewShotImage2019, yangMiningLatentClasses2021, zhangSGOneSimilarityGuidance2020, dingSelfregularizedPrototypicalNetwork2023, chengPOEMPrototypeCross2023}. Despite their efficiency, these methods reduce spatially structured features to single vectors, leading to a loss of fine-grained spatial information~\citep{zhangFewShotSegmentationCycleConsistent2021,liAdaptivePrototypeLearning2021}. To address this, some works introduce multiple prototypes per class
\citep{yangPrototypeMixtureModels2020,liangFoundMissingSemantics2024}. In contrast, matching-based methods \citep{zhangFewShotSegmentationCycleConsistent2021, minHypercorrelationSqueezeFewShot, langLearningWhatNot2022, shiDenseCrossQueryandSupportAttention2022, zhangEfficientSamplingbasedGaussian2025, demarinisLabelAnythingMultiClass2025} compute dense correspondences between support and query images at the pixel level. This preserves spatial granularity and improves performance on segmentation tasks. However, the dense affinity computation incurs higher computational costs, limiting scalability in practical applications.

Despite growing interest in FSS, the field continues to prioritize accuracy over interpretability. The internal mechanisms by which FSS models
arrive at their outputs remain largely unexamined.

\subsection{\texorpdfstring{Interpretability and Explainability}{Interpretability and Explainability} in Few-Shot Learning}

Explainability refers to post-hoc explanations of a model treated as a black box, as opposed to the interpretability of the model itself \citep{rudin2019stop, arrieta2020explainable}. In computer vision, these explanations often take the form of relevance heatmaps, which can be class-specific or class-agnostic. Key explainability categories include saliency-based methods \citep{zhou2018interpreting}, activation-based approaches \citep{erhan2009visualizing, zhang2018top}, perturbation techniques \citep{fong2019understanding}, game-theoretic models like Shapley values \citep{lundberg2017unified}, and attribution methods such as Deep Taylor Decomposition \citep{montavon2017explaining}, LRP \citep{bach2015pixel}, SmoothGrad \citep{smilkov2017smoothgrad}, and Grad-CAM \citep{selvarajuGradCAMVisualExplanations2017}. Nonetheless, their application to FSL remains sparse and under-evaluated.

Some FSL methods embed interpretability directly into their design \citep{wang2021mtunet, li2021scouter, wangMatchThemVisually2023}, while others adapt model-agnostic explainers such as LIME and SHAP \citep{ribeiroWhyShouldTrust2016, acconcjaioco2021one}. These approaches often yield partial explanations confined to specific modules, such as the feature extractor, overlooking inter-module interactions. A separate line of work employs surrogate models (e.g., autoencoders trained on the same data) to generate architecture-independent explanations \citep{utkin2020explanation}, though this introduces dataset dependence and limits generalization.

Direct comparisons between FSL-specific methods and standard model-agnostic tools are largely unexplored; to our knowledge, only one study addresses this comparison \citep{sun2021explanation}, so the field lacks a consistent framework for evaluating explanations in FSL. Model-agnostic explainers also face inherent difficulties in FSL settings involving joint support-query processing: for instance, \citet{fedele2023explain} had to modify a Siamese network's prediction function to integrate LIME, revealing structural incompatibilities between such explainers and FSL models.

Interpretable-by-design classifiers based on prototypical parts, such as ProtoPNet \citep{chen2019lookslikethatdeep}, PIP-Net \citep{Nauta_2023}, and ProtoSeg \citep{sachaProtoSegInterpretableSemantic2023}, with attribution refinements like PRP \citep{gautam2021lookslikethatenhancing} and PixPNet \citep{carmichael2023pixelgroundedprototypicalnetworks}, share with our approach the mechanism of reading latent-space affinities as the explanation: a prediction is explained by the similarity between its features and reference features. They belong, however, to a different family of models, which build interpretability into the architecture through a prototype layer and dedicated training losses. Their prototypes are parameters of a classifier, learned from the training classes, whereas in FSS the class is defined only by the support set given at inference: the reference is not a learned part but the support pixels themselves, and the attribution must cover them. AffEx adds no layer and trains nothing: it reads that attribution from the matching scores a standard matching-based model already computes.

Finally, several few-shot segmentation works visualize support-query correspondences to analyze or motivate their models (e.g., \citep{fan_self-support_2022, boudiaf_few-shot_2021, minHypercorrelationSqueezeFewShot}). These analyses are typically query-centric, focusing on where the query image matches the support examples. In contrast, since our goal is interpretability, we adopt a support-centric perspective: rather than highlighting matches in the query image, we analyze which regions of the support examples contribute to the final segmentation decision. Reading the matching mechanism turns these interactions into an explicit interpretability signal: a practitioner can see which support regions drive a prediction, diagnose a failure, and revise the support set.

\section{Method}
\label{sec:methodology}

\subsection{Matching-Based FSS Models}

\newcommand{\fe}{F}
\newcommand{\support}{\mathcal{S}}

Matching-based Few-Shot Semantic Segmentation models operate by comparing features extracted from support and query images to produce segmentation masks. These models typically consist of a feature extractor $\fe$, a matching mechanism $\text{Match}$, and a segmentation head $\text{Seg}$. Given a support set $\mathcal{S}$ of $K$ image--mask pairs for class $c$, the model predicts a mask $\hat{M}_{q}$ for a query image $I_{q}$ as
\begin{equation}
    \hat{M}_{q} = f_{\theta}(I_{q}, \mathcal{S}) = \text{Seg}_{\theta}\!\big(\text{Match}_{\theta}(\fe_{\theta}(I_{q}), \fe_{\theta}(I_{\support}), M_{\support})\big),
\end{equation}
where $\fe_{\theta}$ is the feature extractor, $\text{Match}_{\theta}$ the matching mechanism, and $\text{Seg}_{\theta}$ the segmentation head, all parameterized by $\theta$. The feature extractor returns a set of multi-level feature maps $\{F^1_\theta(I), \ldots, F^n_\theta(I)\}$, where $F^j_\theta(I) \in \mathbb{R}^{C_j \times H_j \times W_j}$.

The matching mechanism maps support and query features into a common relational space via $\text{Map}^{\text{in}}_\theta$, obtaining $P_\support = \text{Map}^{\text{in}}_\theta(\fe_\theta(I_\support), M_\support)$ and $P_q = \text{Map}^{\text{in}}_\theta(\fe_\theta(I_q))$, and computes similarity scores at each level:
\begin{equation}
    \Sigma = \left\{ \sigma^{j} = \text{MM}_{\theta}(P_{q}^{j}, P_{\support}^{j}) \right\}_{j=1}^{n},\label{eq:scores}
\end{equation}
where $\text{MM}_{\theta}$ denotes the model's matching mechanism, i.e., the operation that compares each query position with each support position in the relational space, e.g., dense cross-attention in DCAMA \citep{shiDenseCrossQueryandSupportAttention2022} or multi-level feature correlation in DMTNet \citep{chenCrossDomainFewShotSemantic2024a}. The scores are then aggregated into an output feature map $F_\text{out} = \text{Map}^{\text{out}}_\theta(\Sigma)$, which the segmentation head processes to produce the final mask $\hat{M}_q = \text{Seg}_\theta(F_\text{out})$. The scores $\sigma^j$ form the direct input to AffEx.

\subsection{Interpreting Matching-Based FSS Models}
\label{sec:interpretation}

The matching scores computed between support and query features constitute an inherent attribution mechanism, one that can be made explicit without requiring gradients or external perturbations. AffEx formalizes this intrinsic property by extracting attribution maps directly from these scores, identifying which support images and features are most influential in determining the segmentation output for a given query image, under the assumption that the matching scores reflect the relevance of each support image to the query.

Let $\mathcal{S}= \{ (I_{i}, M_{i}) \}_{i=1}^{K}$ be the support set with images and corresponding masks. Let $I_{q}\in \mathbb{R}^{C \times H_q \times W_q}$ be the query image, $(x,y)$ a pixel location in $I_{q}$, and $\mathcal{R}\subseteq [1,H_{q}] \times [1,W_{q}]$ a region of interest in $I_{q}$, typically the predicted mask, the ground truth mask, or a user-specified region. We define an attribution function
\begin{equation}
    \label{eq:attribution_function}\mathcal{A}: \left( I_{q}, (x,y), \mathcal{S}\right ) \mapsto \mathbb{R}^{K \times H_s \times W_s},
\end{equation}
which returns, for a given query pixel, a spatial attribution map for each support image. The attribution value at pixel $(x_s, y_s)$ of support image $I_{i}$ is denoted
\begin{equation}
    \alpha_{i,x_s,y_s}= [\mathcal{A}(I_{q}, (x,y), \mathcal{S})]_{i,x_s,y_s}
\end{equation}
and quantifies the contribution of that pixel to the segmentation prediction at $(x,y)$ in the query. The attribution function $\mathcal{A}$ can also be extended to operate over the feature maps of the images: $\mathcal{A}({F_{\theta}}_q, (u,v),{F_\theta}_{\mathcal{S}})$.

We approximate the attribution of a support pixel by that of the corresponding position in the mapped feature map, $\alpha_{i,x_s,y_s}\approx [\mathcal{A}(P_{q}, (u,v), P_{\support})]_{i,u_s,v_s}$ with $(u, v) = \phi(x, y)$ and $(u_s, v_s) = \phi(x_s, y_s)$ the coordinates under the encoder's downsampling, acknowledging that this ignores the encoder's receptive field.

Let $\sigma^{j}\in \mathbb{R}^{H_j \times W_j \times K \times H_j \times W_j}$ be the matching scores for the mapped query feature map $P_{q}^{j}$ and the mapped support feature maps $P_{\support}^{j}$ at the layer $j$ coming from \cref{eq:scores}. These scores serve as the input to the proposed method, illustrated in the \textit{Affinity Explainer} block of \cref{fig:framework}. The attribution function for an element $(u, v)$ in the query feature map can be defined as
\begin{equation}
    \mathcal{A}(P^{j}_{q}, (u,v), P^{j}_{\support}) = \frac{\text{softmax}\left(\sigma^{j}_{u,v}\right)}{\sum_{(u',v')} \text{softmax}\left(\sigma^{j}_{u',v'}\right)} \in \mathbb{R}^{K \times H_j \times W_j},
\end{equation}
where $P^{j}_{q}$ and $P^{j}_{\support}$ are the mapped feature maps for the query and support images at layer $j$, respectively. The softmax is taken over the support dimensions $K \times H_j \times W_j$, i.e., for each query position it normalizes the scores of all support pixels into a probability distribution over the support set. The division is element-wise and the sum runs over all query positions: each support pixel is normalized by the total attention it receives from the query, so that, once averaged over $\mathcal{R}$, its attribution measures the share of that attention coming from $\mathcal{R}$ rather than its overall similarity to the query.

At this step, we obtain a set of attribution maps for each layer \(j\) in the feature extractor. These maps can be aggregated to produce a single attribution map for the query image. The aggregation can be expressed as a weighted combination of the per-layer attributions:
\begin{equation}
    \mathcal{A}(I_{q}, (u, v), \mathcal{S}) = \sum_{j=1}^{n} w_j \cdot \mathcal{A}(P^j_q, (u, v), P^j_{\support}),
\end{equation}
where \(w_j\) are weights reflecting the relative contribution of each layer to the matching process. In our experiments, we use a uniform weighting, i.e., a simple mean of the per-layer attribution maps, which performs on par with weights estimated by feature ablation (\cref{sec:feature_ablation}). Before the mean, each per-layer map, averaged over $\mathcal{R}$, is rescaled to sum to one over the support set, so that no layer dominates through the scale of its values.

In practice, one is interested in the attribution of the entire region $\mathcal{R}$ rather than a single pixel, so we extend the attribution function $\mathcal{A}$ to operate over $\mathcal{R}$ by computing the mean attribution across all its pixels:
\begin{equation}
    \mathcal{A}(I_{q}, \mathcal{R}, \mathcal{S}) := \frac{1}{|\mathcal{R}|}\sum_{(x,y) \in \mathcal{R}}\mathcal{A}(I_{q}, (x,y), \mathcal{S}).
\end{equation}

The attributions computed at this stage already reflect the information conveyed by the support images. The support mask can additionally enter as the sign of the attribution: positive for foreground support pixels, negative for background, after dilating and smoothing the mask to avoid abrupt transitions at its boundaries (ablated in \cref{sec:mask_processing}). We refer to this variant as AffEx, and to the mask-free one as Unmasked AffEx. Which form is the faithful read-out of a model is dictated by where it applies the support mask. If the mask enters \emph{after} the scores AffEx reads (as in DCAMA, whose exposed attention is a pure feature affinity with the mask applied as the attention value), the sign reproduces the model's own next step, turning class-agnostic similarities into a class-specific decision; if the mask is fused \emph{before} matching (as in DMTNet, which masks the support features before correlation), the scores already encode the mask, and applying the sign again is a near no-op.

\subsection{Causal Evaluation Metrics}
\label{sec:metrics}

To evaluate how faithfully an attribution explains the segmentation output, we extend the concepts of the two causal metrics Insertion Area Under the Curve (IAUC) and Deletion Area Under the Curve (DAUC) \citep{petsiukRISERandomizedInput} to the FSS domain. These metrics are designed to quantify the decision perturbation in the model when changing the image according to the explanation.

Defined for classification, these metrics gradually perturb/de-perturb the inputs. To extend them to the FSS scenario, we perturb the support set while leaving the query image unchanged, since our attribution task is based on the support set given the query image.

Given a support set $\mathcal{S}= \{(I_{i}, M_{i})\}_{i=1}^{K}$ and a query image $I_{q}$, let $\Gamma_{t}$ select the top $t$ fraction of support pixels by attribution value, for the attribution map $\mathcal{A}_{\mathcal{R}}= \mathcal{A}(I_{q}, \mathcal{R}, \mathcal{S})$ computed in the region of interest $\mathcal{R}$, let $\Gamma^{c}_{t}$ be its complement, and let $\mathcal{P}(\mathcal{S}, \cdot)$ return the support set with the given pixels perturbed. Sweeping $t$ over a discrete grid $\mathcal{T}$ traces two curves:
\begin{equation}
    \mathrm{IC}_{\mu}(t) = \mu\!\left(f_{\theta}\!\left(I_{q}, \mathcal{P}(\mathcal{S}, \Gamma^{c}_{t}(\mathcal{A}_{\mathcal{R}}))\right)\right),
    \qquad
    \mathrm{DC}_{\mu}(t) = \mu\!\left(f_{\theta}\!\left(I_{q}, \mathcal{P}(\mathcal{S}, \Gamma_{t}(\mathcal{A}_{\mathcal{R}}))\right)\right),
    \label{eq:curves}
\end{equation}
the insertion curve, which restores the most relevant support pixels, and the deletion curve, which removes them. The measure $\mu$ is either the model's confidence over the predicted region $\mathcal{R}_{\hat{M}_q}$ or the mIoU of the prediction against the ground truth; we use both, the second because confidence is uncalibrated and only loosely comparable across models. IAUC and DAUC are the areas under $\mathrm{IC}_{\mu}$ and $\mathrm{DC}_{\mu}$, written IAUC\textsubscript{Conf} and IAUC\textsubscript{mIoU} according to the measure. If the attribution is accurate, deleting the most relevant support pixels degrades the prediction and inserting them restores it.

IAUC and DAUC measure how well a method tracks model behavior, but much of the informative content is concentrated in the initial segment of the curve, where only a small fraction of pixels is perturbed. This is particularly pronounced in the FSS setting, where the model can accurately segment images using only a limited subset of the most relevant pixels. Beyond this region, the curve exhibits largely random fluctuations that contribute little to evaluating explanation quality and instead introduce noise. 

The two metrics we add are read-outs of the same insertion curve, taken where it is informative. IAUC\textsubscript{Conf}@$p$ is its area truncated to $t \leq p$. mIoULoss@$p$ is its deficit at a single early point,
\begin{equation}
    \mathrm{mIoULoss}@p = \mathrm{IC}_{\mathrm{mIoU}}(1) - \mathrm{IC}_{\mathrm{mIoU}}(p),
    \label{eq:miouloss}
\end{equation}
the segmentation accuracy the model loses when it sees only the top $p$ of the support instead of all of it, since at $t{=}1$ nothing is perturbed and $\mathrm{IC}_{\mathrm{mIoU}}(1)$ is the full-support mIoU. Reading the curve at a point is also what makes mIoULoss@$p$ cheap: it costs two forward passes per episode, one on the full support set and one on its top $p$, where a full IAUC or DAUC curve costs one forward pass per perturbation step, 75 in our setting. It also decouples the evaluation from the model's confidence bias, and assesses the explanation directly in terms of segmentation performance.

The two evaluation points are placed by the shape of the insertion curves: $p{=}0.01$ inside the steep initial segment, where explanation methods are most separated, and $p{=}0.05$ at the knee, bracketing the measured points at which the best method reaches 90 percent of its own plateau (1.0 to 6.2 percent of support pixels on DCAMA and DMTNet, in both shot settings). The ranking of the explainers is stable across the whole early grid; the protocol and per-model saturation points are in \cref{sec:p_choice}.

\section{Experiments}
\label{sec:experiments}

We evaluate four matching-based models spanning two generations: DCAMA \citep{shiDenseCrossQueryandSupportAttention2022}, a well-established transformer-based model that uses dense attention as its matching mechanism; DMTNet \citep{chenCrossDomainFewShotSemantic2024a}, a CNN-based model that relies on feature correlation; and two recent pipelines built on vision foundation models: INSID3 \citep{cuttanoINSID3TrainingFreeInContext2026}, a training-free model that drives its prediction with a dense DINOv3 correspondence, and SANSA \citep{cuttano2025sansaunleashinghiddensemantics}, which matches dense SAM2 features. This selection covers four architecturally distinct matching mechanisms. We further analyze two decoder-dominated pipelines, GF-SAM \citep{zhangBridgePointsGraphbased2024} and Matcher \citep{liu2024matchersegmentshotusing}, and a prototype-based head sharing DCAMA's encoder, with details in \cref{sec:decoder_dominated,sec:prototype_head}, respectively. The attributions are evaluated with mIoULoss@$p$, $p \in \{0.01, 0.05\}$, and, in \cref{sec:auc_results}, with the full-curve metrics DAUC\textsubscript{mIoU}, IAUC\textsubscript{mIoU}, their difference, and IAUC\textsubscript{Conf}@0.01. We compare AffEx against several well-established attribution methods applicable to few-shot segmentation: Saliency Maps \citep{simonyan2013deep}, Integrated Gradients \citep{sundararajan2017axiomatic}, Guided Integrated Gradients \citep{kapishnikovGuidedIntegratedGradients2021}, XRAI \citep{kapishnikovXRAIBetterAttributions2019}, Blur IG \citep{xuAttributionScaleSpace2020}, Deep Lift \citep{shrikumarLearningImportantFeatures2017}, and LIME \citep{ribeiroWhyShouldTrust2016}, which we adapt to the support set as described in \cref{sec:lime_fss}. Methods such as Grad-CAM \citep{selvarajuGradCAMVisualExplanations2017} are not directly applicable to few-shot segmentation models. These approaches rely on feature activations from a specific convolutional layer, producing attribution maps with spatial dimensions tied to that layer's output rather than to the support set. As a result, they cannot generate attributions that align spatially with the support images, making them unsuitable for this setting. Also, some methods, such as Guided IG, cannot be applied to specific models (e.g., DMTNet) due to architectural constraints.

In addition to the several established attribution techniques applicable to FSS models, we introduce two simple yet informative baselines. First, a \textit{random} explanation baseline provides a reference point by producing attribution maps entirely independent of the model: it generates a tensor with the same shape as the attribution map, with values sampled uniformly from $[0, 1]$. Second, since FSS models inherently focus on the region defined by the support mask, we define a naive mask-based baseline that uses this prior directly. Specifically, this baseline returns a slightly perturbed version of the support mask as the explanation. We refer to it as the \textit{Gaussian Noise Mask}, computed as $\mathcal{A}_{\mathcal{R}}= M_{s}+ \epsilon$, where $\epsilon \sim \mathcal{N}(0, \sigma^{2})$ and $\sigma = 0.01$. This produces a soft attribution map that reflects the model's spatial prior without relying on learned features.

We evaluated all models on two widely used FSS datasets: COCO $20^{i}$ \citep{nguyen2019feature} and Pascal $5^{i}$ \citep{shabanOneShotLearningSemantic2017}. All experiments were performed on a single NVIDIA A100.

\subsection{Quantitative Evaluation}

\Cref{tab:interpretability_miou} reports the mIoULoss@$p$ results for the 1-way 1-shot and 5-shot settings on both datasets. The proposed AffEx method consistently outperforms all other methods across all four models and datasets: it finds the support pixels that, kept alone, best preserve the prediction. The result carries over unchanged from the classical models to the VFM-based ones: on INSID3 and SANSA, AffEx is best in every setting and beats even the Gaussian Noise Mask baseline. On SANSA, the only VFM-based pipeline where gradient-based attribution is possible, the gradient baselines run but stay behind AffEx. XRAI is the strongest of them and loses in every setting, by between 5.2 and 12.4 mIoULoss points at @0.01. It comes closest at @0.05 with five supports, reaching 3.8 against 3.2 on COCO and 3.9 against 3.8 on Pascal, so a region-based gradient method approaches AffEx when the threshold is loose and the support set is large, while at @0.01 and in the 1-shot setting the gap stays wide. As the mask rule of \cref{sec:interpretation} predicts, the two variants are nearly identical on DMTNet and differ markedly on DCAMA, where the Unmasked variant is nonetheless slightly ahead in two instances, showing that matching scores alone can provide a strong interpretability signal. The strong performance of the Gaussian Noise Mask baseline reflects the dominant role of the support mask structure in guiding model attention, so the support mask alone is already a strong attribution prior. Many established attribution methods fail to recover even this coarse spatial prior, while AffEx goes beyond it. The 1-shot setting confirms the 5-shot ranking at the top, without the exceptions noted above: AffEx is first and its unmasked variant second in every 1-shot column.

\begin{table*}[!tb]
    \centering
    \footnotesize
    \caption{mIoULoss@$p$ results for the evaluation of the attributions on COCO $20^{i}$ and Pascal $5^{i}$ under 1-way 1-shot and 1-way 5-shot settings. The table reports mIoULoss@0.01~$\downarrow$ and mIoULoss@0.05~$\downarrow$. The best results are highlighted in bold, and the second-best results are underlined. Each block lists the methods applicable to that model: gradient-based attribution does not exist on INSID3, whose decision path contains agglomerative clustering and a discrete nearest-neighbor vote, while SANSA, whose adapter-tuned SAM2 pipeline is differentiable end to end, supports it, with the exception of Deep Lift, whose implementation requires the input--baseline pair batched through every module, which SAM2's single-sequence pipeline does not support.}
    \label{tab:interpretability_miou}

    \renewcommand{\arraystretch}{0.92}
    \setlength{\tabcolsep}{4pt}
    
    \begin{tabular*}{\linewidth}{@{\extracolsep{\fill}}lrrrrrrrr}
        \toprule
        & \multicolumn{4}{c}{COCO $20^{i}$} & \multicolumn{4}{c}{Pascal $5^{i}$} \\
        \cmidrule(lr){2-5} \cmidrule(lr){6-9}

Explanation Method & \multicolumn{2}{c}{1-shot} & \multicolumn{2}{c}{5-shot} & \multicolumn{2}{c}{1-shot} & \multicolumn{2}{c}{5-shot} \\
& @0.01 & @0.05 & @0.01 & @0.05 & @0.01 & @0.05 & @0.01 & @0.05 \\
        \midrule
        \multicolumn{9}{c}{DCAMA} \\

Random & 33.79 & 29.78 & 41.42 & 31.62 & 48.72 & 34.64 & 55.24 & 33.42 \\
Gaussian Noise Mask & 27.07 & 12.96 & 31.08 & 17.71 & 33.60 & 16.01 & 35.65 & 15.86 \\
Saliency & 33.93 & 24.13 & 37.01 & 23.73 & 50.06 & 32.80 & 46.42 & 24.46 \\
Integrated Gradients & 32.19 & 22.61 & 34.18 & 22.25 & 44.41 & 27.53 & 45.01 & 23.26 \\
Guided IG & 25.96 & 19.49 & 30.17 & 21.65 & 41.33 & 29.58 & 42.53 & 23.19 \\
Blur IG & 33.89 & 23.29 & 36.69 & 22.84 & 47.96 & 28.61 & 46.86 & 21.53 \\
XRAI & 29.65 & 14.50 & 26.59 & 12.08 & 37.49 & 13.25 & 26.25 & 8.55 \\
Deep Lift & 32.60 & 24.85 & 36.67 & 24.07 & 46.76 & 28.93 & 45.82 & 22.83 \\
LIME & 26.36 & 13.42 & 30.62 & 14.60 & 32.63 & 14.20 & 33.81 & 12.18 \\

Unmasked AffEx (ours) & \underline{19.84} & \underline{10.08} & \underline{17.41} & \textbf{8.85} & \underline{22.85} & \underline{7.90} & \underline{13.78} & \textbf{5.72} \\
AffEx (ours) & \textbf{17.19} & \textbf{7.64} & \textbf{15.38} & \underline{9.83} & \textbf{20.22} & \textbf{6.79} & \textbf{11.56} & \underline{5.85} \\

        \midrule
        \multicolumn{9}{c}{DMTNet} \\

Random & 23.53 & 26.01 & 24.93 & 31.88 & 35.79 & 41.19 & 33.89 & 46.01 \\
Gaussian Noise Mask & 16.20 & 8.66 & 20.69 & 13.26 & 31.02 & 23.73 & 33.24 & 31.00 \\
Saliency & 20.15 & 17.46 & 21.05 & 15.77 & 33.30 & 35.59 & 29.85 & 25.49 \\
Integrated Gradients & 20.53 & 20.81 & 21.63 & 20.32 & 32.03 & 37.64 & 30.31 & 33.07 \\
Blur IG & 20.22 & 17.42 & 21.54 & 20.20 & 31.56 & 35.04 & 29.77 & 33.13 \\
XRAI & 14.83 & 6.30 & 16.48 & 10.73 & 20.56 & 10.80 & 20.03 & 12.88 \\
LIME & 14.07 & 7.85 & 17.91 & 12.48 & 18.90 & 9.57 & 21.14 & 13.83 \\

Unmasked AffEx (ours) & \underline{10.97} & \underline{4.14} & \underline{10.53} & \underline{4.49} & \underline{16.35} & \underline{7.85} & \underline{13.34} & \underline{5.32} \\
AffEx (ours) & \textbf{10.54} & \textbf{3.90} & \textbf{10.35} & \textbf{4.09} & \textbf{16.33} & \textbf{7.83} & \textbf{13.29} & \textbf{5.29} \\

        \midrule
        \multicolumn{9}{c}{INSID3} \\

Random & 59.08 & 46.67 & 64.78 & 63.71 & 60.44 & 47.19 & 68.99 & 68.10 \\
Gaussian Noise Mask & 36.62 & 12.64 & 57.46 & 26.31 & 45.31 & 25.23 & 67.90 & 46.58 \\
LIME & 44.29 & 22.02 & 62.31 & 52.01 & 48.14 & 26.21 & 66.26 & 52.40 \\
Unmasked AffEx (ours) & \underline{26.86} & \underline{4.18} & \underline{28.61} & \underline{7.13} & \underline{33.70} & \underline{4.78} & \underline{44.61} & \underline{10.65} \\
AffEx (ours) & \textbf{23.79} & \textbf{3.05} & \textbf{25.90} & \textbf{6.28} & \textbf{32.99} & \textbf{4.44} & \textbf{43.79} & \textbf{8.92} \\

        \midrule
        \multicolumn{9}{c}{SANSA} \\

Random & 64.54 & 58.87 & 68.50 & 63.42 & 72.55 & 68.27 & 74.26 & 69.83 \\
Gaussian Noise Mask & 41.09 & 17.48 & 53.59 & 15.90 & 56.60 & 28.14 & 66.37 & 25.36 \\
Saliency & 50.77 & 28.23 & 46.69 & 17.16 & 60.25 & 36.78 & 55.97 & 17.00 \\
Integrated Gradients & 52.85 & 30.38 & 49.19 & 22.79 & 61.94 & 36.88 & 61.66 & 28.70 \\
Guided IG & 41.77 & 27.37 & 31.85 & 16.47 & 53.16 & 36.03 & 39.30 & 20.80 \\
Blur IG & 52.79 & 26.17 & 48.56 & 16.45 & 60.07 & 30.83 & 55.49 & 17.10 \\
XRAI & 41.63 & 13.76 & 21.21 & \underline{3.83} & 43.70 & 13.25 & 23.00 & \underline{3.90} \\
LIME & 50.94 & 32.62 & 64.28 & 45.05 & 64.39 & 48.45 & 68.23 & 45.14 \\
Unmasked AffEx (ours) & \underline{35.62} & \underline{13.72} & \underline{18.21} & 4.59 & \underline{40.11} & \underline{12.83} & \underline{17.59} & 4.09 \\
AffEx (ours) & \textbf{29.26} & \textbf{7.48} & \textbf{15.99} & \textbf{3.18} & \textbf{37.86} & \textbf{9.95} & \textbf{16.21} & \textbf{3.83} \\

        \bottomrule
    \end{tabular*}
\end{table*}

The full-curve metrics in \cref{sec:auc_results} agree with this picture: the AffEx variants obtain the highest IAUC\textsubscript{Conf}@0.01 on three of the four models, DMTNet being the exception, and under IAUC $-$ DAUC AffEx is first in five of the eight model and dataset pairs, the other three going to the Gaussian Noise Mask.

We also conducted a computational cost analysis. AffEx runs substantially faster than the integrated-gradient family (182 ms vs.\ 2.6--13 s on DCAMA; 427 ms vs.\ 5--17 s on DMTNet), though not than Saliency and Deep Lift, and uses moderate memory (about 3.3 GB and 2.4 GB in the 1-shot setting). The margin widens on the VFM-based pipelines, which expose a single similarity map, so the read-out adds almost nothing to the forward pass: 1573 ms on INSID3 and 790 ms on SANSA with five supports, against 44 GB and up to 257 s per attribution for the gradient family. Full results are provided in \cref{sec:computational}.

\begin{table}[!tb]
    \centering
    \footnotesize
    \caption{mIoULoss@$p$ results on COCO $20^{i}$ and Pascal $5^{i}$ for the prototype-based head sharing DCAMA's frozen encoder; the results of DCAMA's attention head are in \cref{tab:interpretability_miou}. The table reports mIoULoss@0.01~$\downarrow$ and mIoULoss@0.05~$\downarrow$. The prototype read-out upsamples the pixel-to-prototype similarity, as prototype-part methods do with their activation maps. The best results are highlighted in bold, and the second-best results are underlined.}
    \label{tab:interpretability_proto}

    \renewcommand{\arraystretch}{0.92}
    \setlength{\tabcolsep}{4pt}
    \begin{tabular*}{\linewidth}{@{\extracolsep{\fill}}lrrrrrrrr}
        \toprule
        & \multicolumn{4}{c}{COCO $20^{i}$} & \multicolumn{4}{c}{Pascal $5^{i}$} \\
        \cmidrule(lr){2-5} \cmidrule(lr){6-9}
Explanation Method & \multicolumn{2}{c}{1-shot} & \multicolumn{2}{c}{5-shot} & \multicolumn{2}{c}{1-shot} & \multicolumn{2}{c}{5-shot} \\
& @0.01 & @0.05 & @0.01 & @0.05 & @0.01 & @0.05 & @0.01 & @0.05 \\
        \midrule
Random & 31.13 & 26.54 & 37.17 & 26.55 & 44.52 & 37.29 & 49.76 & 36.12 \\
Gaussian Noise Mask & 24.30 & 14.36 & 29.13 & 18.99 & 39.72 & 22.56 & 42.42 & 19.83 \\
LIME & 20.38 & 10.76 & 25.20 & \textbf{15.91} & 28.09 & 11.28 & 28.89 & \textbf{15.82} \\
Prototype read-out & 16.51 & 10.39 & 19.26 & 19.16 & 24.97 & 11.64 & 20.34 & 19.18 \\
Unmasked AffEx (ours) & \underline{15.61} & \underline{10.07} & \textbf{17.45} & \underline{17.27} & \underline{22.90} & \underline{10.43} & \underline{17.68} & \underline{16.59} \\
AffEx (ours) & \textbf{14.43} & \textbf{9.33} & \underline{17.62} & 18.40 & \textbf{21.84} & \textbf{9.95} & \textbf{17.45} & 17.25 \\
        \bottomrule
    \end{tabular*}
\end{table}

Prototype-based heads are the other classical family of FSS models, so we also test whether the explanation is already inside them. On a PANet-style head \citep{wangPANetFewShotImage2019} sharing DCAMA's frozen encoder, the foreground logit factorizes exactly into a mask-weighted mean of pairwise query--support similarities, so AffEx applies unchanged and subsumes the prototype-style read-out (the upsampled pixel-to-prototype similarity) as its query-blind special case. AffEx and its unmasked variant are the best attributions on this head at the early evaluation point on both datasets and in both shot settings (21.84 at 1-shot on Pascal, against 24.97 for the read-out), and AffEx behaves consistently across the two heads sharing the encoder. \Cref{tab:interpretability_proto} reports the full results; the factorization and the head's training details are given in \cref{sec:panet_details}.

\begin{table}[!tb]
    \centering
    \footnotesize
    \caption{mIoULoss@$p$ results on GF-SAM and Matcher, two decoder-dominated pipelines in which the promptable SAM decoder consumes the whole mask region, in the settings of \cref{tab:interpretability_miou} (1000 episodes each). The table reports mIoULoss@0.01~$\downarrow$ and mIoULoss@0.05~$\downarrow$. Gradient-based baselines are inapplicable, as their decision paths contain discrete, non-differentiable steps (point-graph clustering in GF-SAM, Hungarian matching in Matcher). The best results are highlighted in bold, and the second-best results are underlined.}
    \label{tab:decoder_dominated}

    \renewcommand{\arraystretch}{0.92}
    \setlength{\tabcolsep}{4pt}
    
    \begin{tabular*}{\linewidth}{@{\extracolsep{\fill}}lrrrrrrrr}
        \toprule
        & \multicolumn{4}{c}{COCO $20^{i}$} & \multicolumn{4}{c}{Pascal $5^{i}$} \\
        \cmidrule(lr){2-5} \cmidrule(lr){6-9}
Explanation Method & \multicolumn{2}{c}{1-shot} & \multicolumn{2}{c}{5-shot} & \multicolumn{2}{c}{1-shot} & \multicolumn{2}{c}{5-shot} \\
& @0.01 & @0.05 & @0.01 & @0.05 & @0.01 & @0.05 & @0.01 & @0.05 \\
        \midrule
        \multicolumn{9}{c}{GF-SAM} \\
Random & 46.78 & 14.28 & 39.33 & 5.20 & 40.81 & 14.15 & 36.19 & 5.69 \\
Gaussian Noise Mask & \textbf{15.47} & \textbf{2.98} & \textbf{15.42} & \underline{4.61} & \textbf{11.41} & \textbf{1.74} & \textbf{10.97} & \textbf{3.38} \\
LIME & 42.54 & 27.91 & 59.25 & 48.10 & 44.45 & 26.89 & 62.70 & 53.30 \\
Unmasked AffEx (ours) & 32.24 & 6.74 & 21.52 & \textbf{4.47} & 35.48 & 5.20 & 28.79 & 4.47 \\
AffEx (ours) & \underline{29.63} & \underline{6.20} & \underline{21.15} & 5.27 & \underline{34.66} & \underline{4.77} & \underline{28.17} & \underline{4.40} \\
        \midrule
        \multicolumn{9}{c}{Matcher} \\
Random & 46.16 & 33.92 & 58.04 & 35.29 & 51.49 & 29.86 & 60.08 & 27.58 \\
Gaussian Noise Mask & \textbf{28.81} & \textbf{5.80} & 32.47 & 7.62 & \textbf{28.44} & \textbf{6.05} & \textbf{32.17} & \textbf{5.27} \\
LIME & 40.45 & 26.92 & 52.80 & 41.28 & 43.76 & 32.03 & 56.16 & 42.92 \\
Unmasked AffEx (ours) & 35.81 & 9.66 & \underline{30.31} & \underline{5.93} & 42.68 & 16.29 & 33.62 & 7.41 \\
AffEx (ours) & \underline{35.52} & \underline{7.26} & \textbf{29.54} & \textbf{5.33} & \underline{41.60} & \underline{15.78} & \underline{33.05} & \underline{7.26} \\
        \bottomrule
    \end{tabular*}
\end{table}

Not every VFM-based pipeline is matching-dominated. In GF-SAM \citep{zhangBridgePointsGraphbased2024} and Matcher \citep{liu2024matchersegmentshotusing} the matching does not produce the mask: it seeds point prompts for a promptable SAM decoder, which segments whole objects on its own. On such a \emph{decoder-dominated} pipeline the Gaussian Noise Mask baseline wins the causal metric in seven of the eight settings at @0.01, since the decoder relies on the prompted region as a whole rather than on a discriminative subset of pixels; on Matcher the margin is narrower, and in the COCO 5-shot setting AffEx is best. We read the weakness of AffEx here not as a failure of the read-out but as a measurement: lowering the strongest half of the in-mask similarities at every query position costs GF-SAM and Matcher $31\%$ and $24\%$ of their mIoU, against $58\%$ and $83\%$ for DCAMA and INSID3 (the groups separate only as the deletion deepens: at the top $1\%$, INSID3 loses $10.5\%$ and the other three less than $3\%$), so on these pipelines the decision does not hinge on the strongest matches. The strength of an AffEx attribution is therefore also an estimate of how much the matching mechanism is actually being used, and a low value places the decision in the decoder rather than in the matching. AffEx remains the best model-dependent attribution on both pipelines, as \cref{tab:decoder_dominated} shows; the intervention on the similarities is detailed in \cref{sec:affinity_intervention}.

\subsection{Qualitative Evaluation}

\begin{figure*}[!tb]
    \centering
    \includegraphics[width=\linewidth]{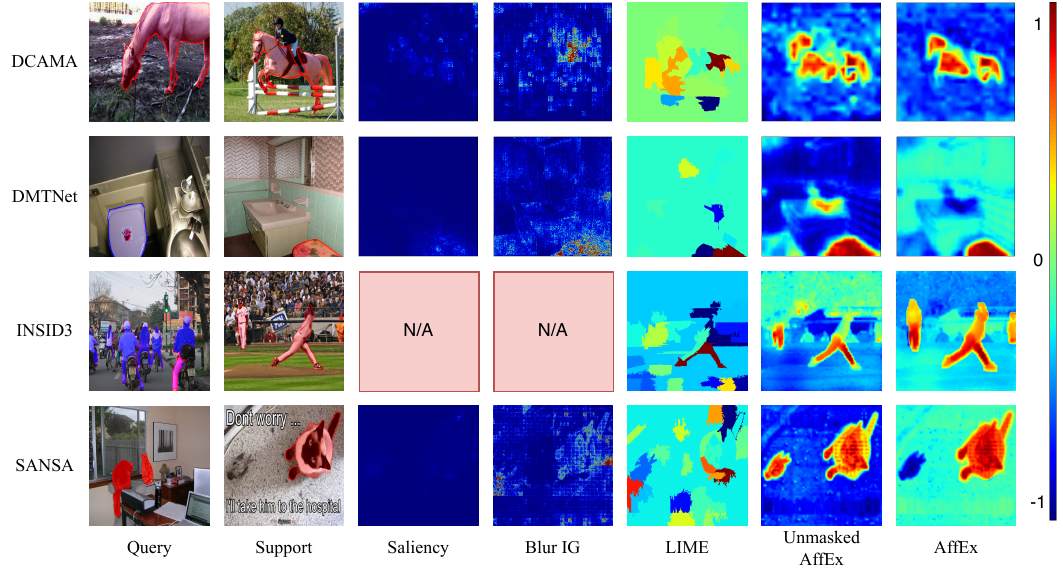}
    \caption{A 1-way 1-shot episode for each of the four evaluated models. The columns show the query and the support image, followed by Saliency, Blur IG, LIME, Unmasked AffEx and AffEx. N/A marks the gradient methods, which do not apply to INSID3. The prediction is highlighted in red over the query image, and the ground truth is shown in blue only when it differs significantly from the prediction. All attribution maps are min-max normalized to $[0,1]$ and drawn with the same colormap.}
    \label{fig:examples}
\end{figure*}

\Cref{fig:examples} presents qualitative results for Saliency, Blur IG, LIME, and the AffEx variants, with one 1-way 1-shot episode for each of the four models. Gradient-based approaches such as Saliency and Blur IG produce noisier, less localized attributions, whereas LIME yields piecewise-constant maps whose shape is set by the superpixel partition of the support image rather than by the model. In contrast, AffEx generates more structured and focused explanations with greater model dependence, and the maps expose how differently the four matching mechanisms resolve the support set: DCAMA highlights distinct regions within the same object, the ResNet features of DMTNet spread over coarse object-level blobs, the DINOv3 correspondence of INSID3 traces the silhouette of the person, and the dense SAM2 features of SANSA cover the whole cat with a sharp outline and a regular grid-like texture.

The INSID3 episode shows how AffEx identifies different levels of importance within a single object: on the support pitcher the attribution peaks on the legs, while the torso and the raised arm receive lower values. The SANSA episode illustrates the mask rule of \cref{sec:interpretation} on semantically related entities. The bird beside the cat is such an entity and lies outside the support mask: Unmasked AffEx gives it a strong positive attribution, since the raw similarity is class-agnostic, while AffEx turns it negative, marking it as background for the queried class, which is the intended, faithful behavior. We also report in \cref{sec:diagnosis} how AffEx supports model diagnosis: on a DCAMA failure case, comparing the attributions computed over the prediction and over the ground truth shows which support region the false positives rest on, and restricting the support mask to the region that matches the query reduces them. More qualitative results are provided in \cref{sec:qualitative_extra}, where the 5-shot examples show a further use of the maps: since AffEx normalizes the attribution jointly over all supports, they also show which shot the prediction rested on. An example of insertion and deletion curves for each model is given in \cref{sec:auc_examples}.

\subsection{Ablation Study}

We adopt uniform averaging across feature levels as the aggregation strategy, as more sophisticated mechanisms such as feature ablation do not yield significant performance gains (see \cref{sec:feature_ablation}). The softmax normalizes raw matching scores into a probability distribution, so their relative contributions to the query prediction can be compared. Removing it leads to a degradation of 12.58 and 5.09 mIoULoss@$p$ points for $p \in \{0.01, 0.05\}$ on DCAMA in a 1-way 5-shot setting on Pascal $5^{i}$; no degradation is observed with DMTNet, as its correlation-based matching is inherently normalized. The role of the support mask is ablated through the Unmasked AffEx variant reported in \cref{tab:interpretability_miou} and \cref{tab:interpretability_auc}. Since the two variants differ by both the sign and the mask smoothing, a smoothed-unmasked control (same blur, no mask, no sign; DCAMA, Pascal $5^{i}$, 5-shot, mIoULoss@0.01) scores 12.12, between Unmasked AffEx (13.78) and AffEx (11.56): the smoothing accounts for part of the gap, but the sign is required for the best result. Additional ablation studies cover feature aggregation (\cref{sec:feature_ablation}), perturbation steps (\cref{sec:steps_ablation}), and support mask processing (\cref{sec:mask_processing}).

\section{Conclusion}
\label{sec:conclusion}

We introduced support-side attribution for few-shot segmentation and AffEx, which solves this task by reading the matching scores the models already compute, without gradients or external perturbations, together with a causal evaluation framework based on support-set perturbation. The read-out consumes nothing the models do not already compute, which is the sense in which they are interpretable by design. Across an extensive benchmark, AffEx consistently outperforms existing explainability techniques, quantitatively and qualitatively, from classical matching-based models to vision-foundation-model pipelines, where the framework further diagnoses whether a model's decision is carried by its matching stage or reshaped by its decoder, and on a prototype-based head, where AffEx subsumes the prototype-style read-out as a special case. Future directions include combining AffEx with decoder-side explainers and refining its maps with relevance propagation.

\paragraph{Limitations.} AffEx explains the matching stage: on decoder-dominated pipelines such as GF-SAM and Matcher, what the decoder adds to the prediction lies outside the attribution. Mapping each support pixel to its position on the feature grid ignores the encoder's receptive field. The causal metrics replace support pixels with zeros, so the model is evaluated on inputs unlike those seen in training, and all evaluations are 1-way, leaving multi-way episodes, where support classes compete, untested.

\begin{ack}

We acknowledge ISCRA for awarding this project access to the LEONARDO supercomputer, owned by the EuroHPC Joint Undertaking, hosted by CINECA (Italy). The authors declare no competing interests.
\end{ack}

\bibliographystyle{plainnat}
\bibliography{main}

\clearpage
\appendix

\section{Considerations on Computational Cost}
\label{sec:computational}

This section analyzes the computational costs of our proposed Affinity Explainer in comparison with widely used explanation techniques: Saliency Maps \citep{simonyan2013deep}, Integrated Gradients \citep{sundararajan2017axiomatic}, Guided Integrated Gradients \citep{kapishnikovGuidedIntegratedGradients2021}, XRAI \citep{kapishnikovXRAIBetterAttributions2019}, Blur IG \citep{xuAttributionScaleSpace2020}, Deep Lift \cite{shrikumarLearningImportantFeatures2017}, and LIME \citep{ribeiroWhyShouldTrust2016}.

We evaluated both the memory consumption and inference time required to produce explanations for two state-of-the-art few-shot segmentation models, DCAMA \cite{shiDenseCrossQueryandSupportAttention2022} and DMTNet \cite{chenCrossDomainFewShotSemantic2024a}, and for the two VFM-based pipelines, INSID3 \cite{cuttanoINSID3TrainingFreeInContext2026} and SANSA \cite{cuttano2025sansaunleashinghiddensemantics}, under 1-shot and 5-shot settings. All experiments were conducted on an NVIDIA A100 GPU. AffEx interpolates the per-layer similarity maps to a common resolution before aggregation, which sets both its memory footprint and part of its latency. We used the native resolution of the backbone for DCAMA and DMTNet, and $64{\times}64$ for INSID3 and SANSA, whose inputs are $1024{\times}1024$ and for which the native resolution is not affordable. On DCAMA and DMTNet, which expose many per-layer maps (22 on DCAMA, one per Swin layer from the second stage on), this read-out accounts for the time beyond the forward pass.

As reported in \cref{tab:computational_cost}, AffEx is not as lightweight as Saliency Maps but is an order of magnitude faster than Integrated Gradients and its variants. This efficiency stems from AffEx's design, which reads similarity maps the model has already computed and avoids backward passes through the network. In terms of memory usage, AffEx maintains a moderate footprint, requiring less memory than Deep Lift, which is the most memory-intensive method evaluated. The gap widens on the VFM-based pipelines. INSID3 and SANSA expose a single similarity map, so the read-out adds almost nothing and an explanation costs the forward pass itself: 765~ms against 764~ms on INSID3 and 282~ms against 282~ms on SANSA in the 1-shot setting. The baselines pay the usual multiplier, with LIME at 18.9~s on INSID3 and Integrated Gradients at 16.2~s on SANSA, and the gradient family also becomes memory bound on SANSA, where a single Saliency map needs 15.43~GB with one support and 44.08~GB with five, against 2.43~GB and 5.40~GB for AffEx.

AffEx needs a single forward pass and no backward pass, which keeps it usable where the gradient family is not, such as on a GPU that cannot hold 44~GB.

\begin{table*}
    [tb]
    \centering
    \caption{Computational cost comparison between AffEx and other explanation methods, including Integrated Gradients, Saliency Maps, and Deep Lift. The table reports the average memory consumption (in GB) and time (in milliseconds) for generating explanations, along with the forward pass time of the models for reference. The forward pass cost is shown in a gray background. The results are presented for both 1-shot and 5-shot scenarios, with standard deviations included. The experiments were conducted on an NVIDIA A100. For INSID3 and SANSA the AffEx rows are measured at similarity map resolution $64{\times}64$, while the gradient-based and perturbation-based baselines operate at the native input resolution of each model.}
    \label{tab:computational_cost}
    \resizebox{\linewidth}{!}{
    \begin{tabular}{@{}llrrrr@{}}
        \toprule \multicolumn{1}{l}{\textbf{}}                          & \multicolumn{1}{l}{}                   & \multicolumn{2}{c}{1 shot}          & \multicolumn{2}{c}{5 shots}          \\
        \cmidrule(lr){3-4} \cmidrule(lr){5-6} \multicolumn{1}{c}{Model} & \multicolumn{1}{c}{Explanation Method} & \multicolumn{1}{c}{Expl. Mem. (GB)} & \multicolumn{1}{c}{Expl. Time (ms)} & \multicolumn{1}{c}{Expl. Mem. (GB)} & \multicolumn{1}{c}{Expl. Time (ms)} \\
        \midrule 
        \multirow{9}{*}{DCAMA}                                 
        & \forward{Forward Pass}                 & \forward{1.85 ± 0.00}               & \forward{44 ± 1}                    & \forward{6.37 ± 0.00}               & \forward{109 ± 1}                   \\
                                                                        & Saliency                               & 4.00 ± 0.00                         & 102 ± 1                             & 13.70 ± 0.00                        & 284 ± 2                             \\
                                                                        & Integrated Gradients                   & 4.01 ± 0.00                         & 2571 ± 40                           & 13.76 ± 0.00                        & 7154 ± 103                          \\
                                                                        & Guided IG                              & 4.00 ± 0.00                         & 3693 ± 34                           & 13.70 ± 0.00                        & 11079 ± 78                          \\
                                                                        & Blur IG                                & 4.00 ± 0.00                         & 11904 ± 56                          & 13.70 ± 0.00                        & 35268 ± 163                         \\
                                                                        & XRAI                                   & 4.00 ± 0.00                         & 13076 ± 1469                        & 13.70 ± 0.00                        & 37647 ± 3645                        \\
                                                                        & Deep Lift                              & 6.93 ± 0.00                         & 146 ± 1                             & 24.09 ± 0.01                        & 442 ± 2                             \\
                                                                        & LIME                                   & 1.85 ± 0.00                         & 1432 ± 91                           & 6.30 ± 0.00                         & 3627 ± 98                           \\
                                                                        & AffEx (ours)                           & 3.33 ± 0.21                         & 182 ± 5                             & 13.90 ± 1.03                        & 552 ± 28                            \\
        \midrule \multirow{8}{*}{DMTNet}                                & \forward{Forward Pass}                 & \forward{0.83 ± 0.00}               & \forward{142 ± 1}                   & \forward{2.67 ± 0.00}               & \forward{667 ± 19}                  \\
                                                                        & Saliency                               & 6.49 ± 0.00                         & 206 ± 2                             & 20.17 ± 0.00                        & 1039 ± 90                           \\
                                                                        & Integrated Gradients                   & 6.51 ± 0.00                         & 5141 ± 135                          & 20.23 ± 0.00                        & 27096 ± 375                         \\
                                                                        & Blur IG                                & 6.45 ± 0.00                         & 15153 ± 114                         & 19.97 ± 0.00                        & 55707 ± 410                         \\
                                                                        & XRAI                                   & 6.45 ± 0.00                         & 16735 ± 1569                        & 19.97 ± 0.00                        & 58245 ± 4145                        \\
                                                                        & Deep Lift                              & 12.92 ± 0.00                        & 279 ± 2                             & -                                   & -                                   \\
                                                                        & LIME                                   & 0.85 ± 0.00                         & 5087 ± 93                           & 2.76 ± 0.00                         & 22717 ± 1039                        \\
                                                                        & AffEx (ours)                           & 2.38 ± 0.22                         & 427 ± 5                             & 11.13 ± 0.84                        & 2072 ± 20                           \\
        \midrule \multirow{3}{*}{INSID3}                          & \forward{Forward Pass}                 & \forward{2.04 ± 0.00}               & \forward{764 ± 12}                  & \forward{3.19 ± 0.00}               & \forward{1568 ± 16}                 \\
                                                                        & LIME                             & 2.24 ± 0.00                         & 18883 ± 848                         & 3.84 ± 0.00                         & 42142 ± 1080                        \\
                                                                        & AffEx (ours)                     & 2.04 ± 0.00                         & 765 ± 12                            & 3.19 ± 0.00                         & 1573 ± 13                           \\
        \midrule \multirow{8}{*}{SANSA}                           & \forward{Forward Pass}                 & \forward{2.43 ± 0.00}               & \forward{282 ± 0}                   & \forward{5.40 ± 0.00}               & \forward{787 ± 0}                   \\
                                                                        & Saliency                         & 15.43 ± 0.00                        & 645 ± 1                             & 44.08 ± 0.00                        & 1808 ± 2                            \\
                                                                        & Integrated Gradients             & 15.54 ± 0.00                        & 16204 ± 24                          & 44.42 ± 0.00                        & 45310 ± 37                          \\
                                                                        & Guided IG                        & 15.47 ± 0.00                        & 27817 ± 781                         & 44.20 ± 0.00                        & 87119 ± 1498                        \\
                                                                        & Blur IG                          & 15.47 ± 0.00                        & 84539 ± 358                         & 44.20 ± 0.00                        & 256821 ± 333                        \\
                                                                        & XRAI                             & 15.47 ± 0.00                        & 51550 ± 6354                        & 44.20 ± 0.00                        & 221336 ± 22408                      \\
                                                                        & LIME                             & 2.63 ± 0.00                         & 7794 ± 22                           & 6.05 ± 0.00                         & 23360 ± 580                         \\
                                                                        & AffEx (ours)                     & 2.43 ± 0.00                         & 282 ± 0                             & 5.40 ± 0.00                         & 790 ± 0                             \\
        \bottomrule
    \end{tabular}}
\end{table*}

\section{Adapting LIME to Few-Shot Segmentation}
\label{sec:lime_fss}
While gradient-based and perturbation-based attribution methods can be directly applied to FSS models with minor modifications, adapting model-agnostic approaches such as LIME \cite{ribeiroWhyShouldTrust2016} requires a more careful formulation. We adapted the LIME framework to the FSS setting, where predictions depend jointly on a query image and a structured support set. The interpretable feature space is defined as a binary activation vector over the indices of the $L$ superpixels into which each support image is partitioned (see below): entry $l$ indicates whether the $l$-th superpixel of every support image, together with the corresponding region of its mask, is kept in the conditioning set, while the query remains fixed. Perturbations are generated by switching entries off at random following a Bernoulli process, with removed regions replaced by zero-valued baselines; a support whose mask becomes empty is dropped. The model is then evaluated on each perturbed configuration to fit a local linear surrogate model around the original prediction.

To keep the explanation spatially coherent, each support image was partitioned into about $L$ superpixels using the SLIC algorithm \cite{achantaSLICSuperpixelsCompared2012}, which provides the feature mask for LIME. Each superpixel index is an interpretable unit, switched on or off as a whole during perturbation, ensuring that the explanation remains coherent with perceptual boundaries.

Locality between the original support set $\mathcal{S}=\{(I_{i},M_{i})\}_{i=1}^{K}$ and a perturbed configuration $\mathcal{S}'$ was enforced through a composite distance:
\begin{equation}
    D(\mathcal{S},\mathcal{S}') = \frac{1}{K}\!\sum_{i=1}^{K}\!\big[ \lambda(1-\cos(I_{i},I_{i}')) + (1-\lambda)\,\Delta_{\text{fg}}(M_{i},M_{i}')\big],
\end{equation}
where $\cos(\cdot,\cdot)$ measures cosine similarity between support images and perturbed support images, and $\Delta_{\text{fg}}$ quantifies the fraction of foreground pixels removed in each perturbed mask. The hyperparameter $\lambda \in [0, 1]$ serves as a balancing weight to adjust the relative influence of the visual dissimilarity (image-based) and the mask dissimilarity (mask-based) components of the overall distance. Missing supports are assigned maximal dissimilarity ($D_{i}=1$). The corresponding kernel weight is
\begin{equation}
    w(\mathcal{S},\mathcal{S}') = \exp\!\left(-\tfrac{D(\mathcal{S},\mathcal{S}')^2}{\sigma^2}\right),
\end{equation}
with $\sigma$ controlling the width of the locality kernel. This adaptation allows LIME to account for the dual conditioning structure of FSS, producing explanations that remain localized in both spatial and semantic terms. To find the optimal number of superpixels $L$, $\lambda$, and $\sigma$, we performed a grid search on Pascal $5^i$ in a 5-shot setting, using the DCAMA model, optimizing for mIoULoss@0.01. Results are reported in \cref{tab:lime_fss_params}. We report results only for the best $\lambda$ as it had a marginal effect on performance. The best configuration uses $L{=}160$ superpixels, Ridge regression, $\sigma{=}0.5$, and $\lambda{=}0.2$.

\begin{table}[!t]
    \centering
    \small
    \setlength{\tabcolsep}{5pt}
    \renewcommand{\arraystretch}{1.1}
    \caption{Grid search results for adapting LIME to few-shot segmentation. The table reports the mIoULoss@0.01 for different configurations of superpixels, regression models (Lasso or Ridge), kernel width ($\sigma$), and the balance parameter ($\lambda$). The best performance is highlighted in bold. }
    \label{tab:lime_fss_params}
    \begin{tabular}{c c c c c}
        \toprule Superpixels          & Model                  & Kernel           & Image-Mask          & mIoUL.         \\
                                      &                        & width ($\sigma$) & balance ($\lambda$) & @0.01          \\
        \midrule \multirow{9}{*}{80}  & \multirow{3}{*}{Lasso} & 0.5              & 0.2                 & 37.54          \\
                                      &                        & 0.5              & 0.5                 & 37.65          \\
                                      &                        & 0.5              & 0.8                 & 37.39          \\[2pt]
                                      & \multirow{3}{*}{Lasso} & 1.0              & 0.2                 & 37.24          \\
                                      &                        & 1.0              & 0.5                 & 37.19          \\
                                      &                        & 1.0              & 0.8                 & 37.23          \\[2pt]
                                      & \multirow{3}{*}{Ridge} & 0.5              & 0.2                 & 36.31          \\
                                      &                        & 0.5              & 0.8                 & 36.83          \\
                                      &                        & 3.0              & 0.5                 & 36.46          \\
        \midrule \multirow{9}{*}{160} & \multirow{3}{*}{Lasso} & 0.5              & 0.2                 & 37.47          \\
                                      &                        & 0.5              & 0.5                 & 37.16          \\
                                      &                        & 0.5              & 0.8                 & 36.63          \\[2pt]
                                      & \multirow{3}{*}{Lasso} & 1.0              & 0.2                 & 36.36          \\
                                      &                        & 1.0              & 0.5                 & 36.54          \\
                                      &                        & 1.0              & 0.8                 & 36.52          \\[2pt]
                                      & \multirow{3}{*}{Ridge} & 0.5              & 0.2                 & \textbf{33.82} \\
                                      &                        & 0.5              & 0.8                 & 33.97          \\
                                      &                        & 3.0              & 0.5                 & 34.05          \\
        \bottomrule
    \end{tabular}
\end{table}

\section{Feature Ablation}
\label{sec:feature_ablation}

In addition to the simple mean aggregation of per-layer attribution maps, we explored an alternative aggregation method, \emph{Feature Ablation}. In general, the aggregation can be expressed as a weighted combination of the per-layer attributions:
\begin{equation}
    \mathcal{A}(I_{q}, (u, v), \mathcal{S}) = \sum_{j=1}^{n} w_j \cdot \mathcal{A}(P^j_q, (u, v), P^j_{\support}),
\end{equation}
where \(w_j\) reflects the contribution of layer \(j\) to the matching process. Feature Ablation estimates these weights by evaluating the impact of removing individual layers on the overall attribution, highlighting the layers that contribute most significantly.  

We found that replacing the ablation-based weights with a uniform mean did not result in a significant performance change. While each model could exhibit biases toward certain feature levels, averaging provides a sufficiently accurate approximation of per-layer contributions. Accordingly, we employed a simple mean for aggregation in our main experiments.

\section{Full-Curve Insertion and Deletion Results}
\label{sec:auc_results}

\begin{table}[!tb]
    \centering
    \scriptsize
    \caption{Insertion and deletion evaluation under 1-way 5-shot. The table reports IAUC~$\uparrow$, deletion DAUC~$\downarrow$, their difference (Diff $\uparrow$), and the insertion AUC computed on the prediction confidence over the first 1\% of revealed pixels (IAUC\textsubscript{Conf}@0.01 $\uparrow$). No value is highlighted here: the full-curve IAUC/DAUC are dominated by the flat tail of the curves, whose oscillations are of the same order as the differences between methods, so their ranking is not robust; Diff (IAUC $-$ DAUC) summarizes the separation between informative and non-informative regions and is reported for completeness. On INSID3 gradient-based attribution does not exist, as its decision path contains discrete steps. On SANSA it does, but the cost of the gradient family is prohibitive at this scale, so we consider here only XRAI, the strongest gradient baseline on SANSA under mIoULoss@$p$ in \cref{tab:interpretability_miou}, where the whole family is reported.}
    \label{tab:interpretability_auc}

    \begin{tabularx}
        {\textwidth}{l *{8}{>{\raggedleft\arraybackslash}X}} \toprule & \multicolumn{4}{c}{COCO $20^{i}$} & \multicolumn{4}{c}{Pascal $5^{i}$} \\
        \cmidrule(lr){2-5}\cmidrule(lr){6-9}

        Explanation Method & \makecell{IAUC\\mIoU} & \makecell{DAUC\\mIoU} & Diff & \makecell{IAUC\\Conf\\@0.01} & \makecell{IAUC\\mIoU} & \makecell{DAUC\\mIoU} & Diff & \makecell{IAUC\\Conf\\@0.01} \\
        \midrule \multicolumn{9}{c}{DCAMA} \\
        Random & 47.50 & 46.85 & 0.65 & 43.36 & 71.35 & 70.77 & 0.59 & 41.71 \\
        Gaussian Noise Mask & 52.32 & 15.46 & 36.86 & 53.80 & 74.67 & 23.36 & 51.31 & 55.32 \\
        Saliency & 52.02 & 41.10 & 10.92 & 44.97 & 72.99 & 66.73 & 6.27 & 44.32 \\
        Integrated Gradients & 49.23 & 45.97 & 3.27 & 45.63 & 72.38 & 70.22 & 2.16 & 42.94 \\
        Guided IG & 50.77 & 45.10 & 5.67 & 51.25 & 71.74 & 70.35 & 1.39 & 49.08 \\
        Blur IG & 52.40 & 41.63 & 10.78 & 44.25 & 73.82 & 65.80 & 8.01 & 42.23 \\
        XRAI & 54.45 & 38.72 & 15.73 & 51.98 & 75.52 & 60.98 & 14.55 & 56.12 \\
        Deep Lift & 49.78 & 48.51 & 1.27 & 44.80 & 72.72 & 71.66 & 1.06 & 44.19 \\
        LIME & 53.18 & 48.44 & 4.73 & 47.98 & 74.77 & 72.20 & 2.57 & 49.19 \\
        Unmasked AffEx (ours) & 54.85 & 28.26 & 26.59 & 60.82 & 76.23 & 46.34 & 29.89 & 66.82 \\
        AffEx (ours) & 54.86 & 16.30 & 38.57 & 60.49 & 75.92 & 25.41 & 50.51 & 67.39 \\
        \midrule \multicolumn{9}{c}{DMTNet} \\
        Random & 21.21 & 20.98 & 0.23 & 60.36 & 29.01 & 28.46 & 0.55 & 60.35 \\
        Gaussian Noise Mask & 39.08 & 15.71 & 23.37 & 59.35 & 54.53 & 24.41 & 30.12 & 59.90 \\
        Saliency & 36.51 & 26.79 & 9.72 & 63.09 & 52.43 & 41.22 & 11.21 & 64.85 \\
        Integrated Gradients & 26.25 & 19.47 & 6.78 & 62.51 & 37.13 & 29.08 & 8.05 & 63.99 \\
        Blur IG & 35.04 & 20.32 & 14.72 & 62.17 & 47.83 & 33.82 & 14.01 & 63.62 \\
        XRAI & 38.19 & 26.55 & 11.64 & 68.42 & 55.82 & 45.72 & 10.09 & 69.78 \\
        LIME & 36.83 & 32.44 & 4.39 & 70.72 & 54.76 & 49.80 & 4.96 & 70.56 \\
        Unmasked AffEx (ours) & 40.10 & 17.09 & 23.01 & 67.30 & 58.16 & 28.81 & 29.35 & 66.80 \\
        AffEx (ours) & 40.53 & 16.53 & 24.00 & 67.10 & 58.77 & 27.34 & 31.43 & 66.77 \\
        \midrule \multicolumn{9}{c}{INSID3} \\
        Random & 38.82 & 38.14 & 0.68 & 26.89 & 36.06 & 35.19 & 0.88 & 26.89 \\
        Gaussian Noise Mask & 61.73 & 3.95 & 57.79 & 30.38 & 64.02 & 5.82 & 58.20 & 28.42 \\
        LIME & 49.42 & 38.69 & 10.73 & 27.47 & 52.93 & 41.88 & 11.05 & 27.41 \\
        Unmasked AffEx (ours) & 63.72 & 5.91 & 57.81 & 39.88 & 67.50 & 7.67 & 59.83 & 35.02 \\
        AffEx (ours) & 63.14 & 4.37 & 58.77 & 41.19 & 68.10 & 7.97 & 60.13 & 35.32 \\
        \midrule \multicolumn{9}{c}{SANSA} \\
        Random & 59.13 & 58.30 & 0.83 & 31.06 & 59.98 & 59.02 & 0.95 & 28.90 \\
        Gaussian Noise Mask & 72.87 & 15.90 & 56.97 & 37.26 & 74.65 & 18.20 & 56.46 & 32.28 \\
        XRAI & 74.82 & 42.39 & 32.43 & 48.11 & 77.16 & 52.73 & 24.43 & 47.78 \\
        LIME & 70.55 & 68.21 & 2.33 & 32.75 & 72.82 & 71.39 & 1.43 & 31.25 \\
        Unmasked AffEx (ours) & 74.60 & 20.83 & 53.76 & 51.99 & 76.95 & 25.73 & 51.22 & 52.25 \\
        AffEx (ours) & 74.60 & 18.27 & 56.33 & 53.66 & 77.04 & 29.72 & 47.32 & 53.19 \\
        \bottomrule
    \end{tabularx}
\end{table}

\Cref{tab:interpretability_auc} reports the results of the evaluation of the attributions on COCO $20^{i}$ and Pascal $5^{i}$ for the four evaluated models in a 1-way 5-shot setting, in terms of IAUC and DAUC. The IAUC\textsubscript{mIoU} and DAUC\textsubscript{mIoU} values vary across models, including the random baseline, because they reflect each model's intrinsic performance. Consequently, IAUC and DAUC alone are not directly comparable across models.

To provide a more meaningful assessment, we compute the difference $\text{IAUC} - \text{DAUC}$ for each explanation method and model. This difference, therefore, reflects the separation between informative and non-informative regions identified by the explanation method. Higher values indicate more effective attribution, whereas values close to zero suggest that the identified regions have a limited impact.

However, this measure should be interpreted cautiously: $\text{IAUC} - \text{DAUC}$ is a coarse-grained metric influenced by fluctuations in the latter portions of the curves, which have a limited impact on model performance. Despite this, the substantial performance gap between the Gaussian Noise Mask and the other competitors shows that these methods do not capture the structural foreground prior that FSS models inherently exploit. We also observe that the Unmasked variant exhibits a poorer DAUC. Lacking class-specific information from the support mask, it identifies pixels that are relevant to the model but not necessarily to the target class, resulting in lower DAUC. Nevertheless, even without the support mask, the method can still identify regions informative for the target class. Explanation quality therefore depends on the model's priors and on how well the method uses them.

We also report the IAUC\textsubscript{Conf}@0.01 results, which measure the model's confidence increase when the top 1\% of pixels are added to the support set. The AffEx variants achieve the highest IAUC\textsubscript{Conf}@0.01 for DCAMA, while for DMTNet LIME and XRAI perform slightly better. On the two VFM-based pipelines AffEx leads by 10.8 points on INSID3 (41.19 against 30.38 for the Gaussian Noise Mask) and by 5.6 on SANSA (53.66 against 48.11 for XRAI), over COCO $20^{i}$, with the same picture on Pascal $5^{i}$. The Diff column ranks differently on SANSA, where the Gaussian Noise Mask separates the two curves more than AffEx does, because IAUC $-$ DAUC is dominated by the flat tail of the curves: the two methods are 2.4 IAUC points apart while their DAUC differs by more than 10, and the ordering under mIoULoss@$p$ in \cref{tab:interpretability_miou} and under IAUC\textsubscript{Conf}@0.01 is unambiguous in favor of AffEx.

\section{Ablation on the Number of Steps for IAUC and DAUC}
\label{sec:steps_ablation}

IAUC and DAUC are metrics that approximate an area under the curve. Thus, they require several steps to compute the area. Each of these steps requires evaluating the model, which can be expensive depending on the model used. Furthermore, the model's confidence may fluctuate when adding or removing pixels, making it difficult to determine the correct approximation. To address this issue, we performed an ablation study on the number of steps used to compute the IAUC and DAUC metrics. We varied the number of steps from 5 to 1000, using the value obtained with 1000 steps as the ground truth, and computed the relative error with respect to it. We evaluated this ablation on the Pascal~$5^{i}$ dataset, using the DCAMA model and the Affinity Explainer and Saliency methods. We calculated the 95th percentile to find an upper bound of the relative error. \Cref{tab:steps_ablation} shows the results of the ablation study on the number of steps used to compute the IAUC and DAUC metrics. As the number of steps increases, the relative error decreases, indicating that the metrics become more stable. As a trade-off between computational cost and stability, we suggest 75 steps, which we used in all the experiments of this paper.

\begin{table}[!tb]
    \centering
    \footnotesize
    \caption{Relative error on the 95th percentile of the Insertion Area Under the Curve (IAUC) and Deletion Area Under the Curve (DAUC) metrics with respect to the ground truth value obtained with 1000 steps. The table shows the relative error for the Affinity Explainer (AffEx) and Saliency methods. The lower the relative error, the more stable the metric is with respect to the number of steps.}
    \label{tab:steps_ablation}
    \begin{tabular}{llrrrrrr}
        \toprule                       & N. Steps & 5    & 10   & 25   & 50   & 75   & 100  \\
        Metric                         & Method   &      &      &      &      &      &      \\
        \midrule \multirow{2}{*}{DAUC} & Saliency & 6.44 & 3.85 & 1.91 & 1.16 & 0.86 & 0.66 \\
                                       & AffEx    & 9.04 & 5.37 & 3.12 & 1.85 & 1.37 & 1.10 \\
        \multirow{2}{*}{IAUC}          & Saliency & 6.33 & 3.63 & 2.14 & 1.32 & 0.98 & 0.75 \\
                                       & AffEx    & 6.64 & 3.22 & 1.34 & 0.77 & 0.59 & 0.46 \\
        \bottomrule
    \end{tabular}
\end{table}

\section{Choice of the Evaluation Points $p$}
\label{sec:p_choice}

\begin{figure}[!tb]
    \centering
    \includegraphics[width=.75\linewidth]{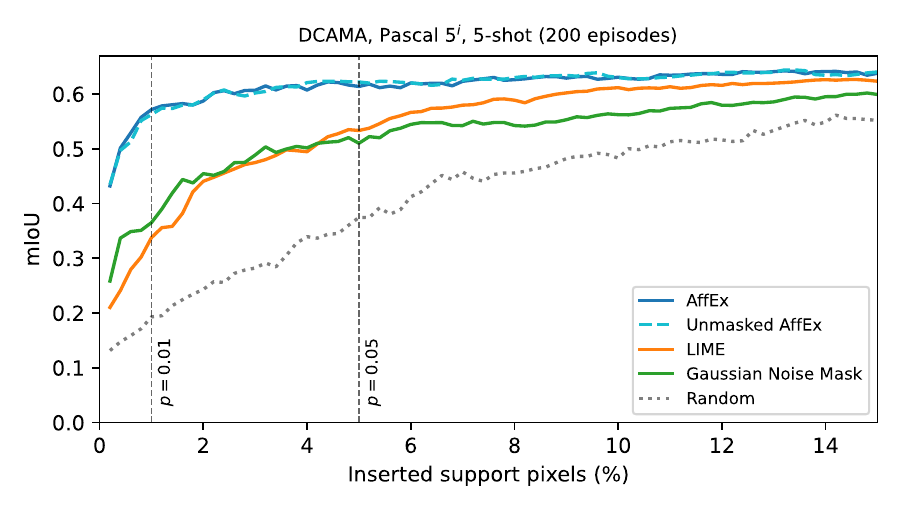}
    \caption{Insertion curves at fine resolution over the early regime (steps of 0.2 percent of support pixels) for DCAMA on Pascal $5^{i}$, 5-shot, averaged over episodes. The curves of the strong explainers saturate within the first few percent; $p{=}0.01$ falls in the steep segment where methods are most separated, $p{=}0.05$ at the knee of the best curves.}
    \label{fig:insertion_curves}
\end{figure}

The two evaluation points of mIoULoss@$p$ are placed by the shape of the insertion curves: one inside the steep initial segment, where explanation methods are most separated, and one around the point where the best curves have flattened. To back this placement quantitatively, we computed insertion curves for DCAMA and DMTNet at fine resolution over the early regime: 75 points over the first 15 percent of support pixels, i.e., steps of 0.2 percent, on Pascal $5^{i}$, in both 1-shot and 5-shot settings. \Cref{fig:insertion_curves} shows the resulting curves for DCAMA in the 5-shot setting. For each explainer, we measured the smallest $p$ at which its episode-mean curve reaches 90 and 95 percent of its own plateau (the curve value at 15 percent); \cref{tab:saturation} reports these saturation points.

\begin{table}[!tb]
    \centering
    \caption{Saturation points of the insertion curves: smallest $p$ (in percent of support pixels) at which each explainer's episode-mean insertion curve reaches 90\,/\,95 percent of its own plateau. Pascal $5^{i}$, fine grid of 0.2 percent steps. Random on DMTNet is non-monotone (inserting randomly chosen support pixels lowers the mIoU below its own starting value), so its saturation point is undefined. LIME's early points on DMTNet come with a visibly lower plateau (0.42 against 0.48 for AffEx at 1-shot): saturating early to a worse value is not a sign of informativeness.}
    \label{tab:saturation}
    \small
    \begin{tabular}{lcccc}
        \toprule
        Explainer & DCAMA 1-shot & DCAMA 5-shot & DMTNet 1-shot & DMTNet 5-shot \\
        \midrule
        AffEx & 3.2 / 5.6 & 1.0 / 2.2 & 6.2 / 9.0 & 4.8 / 8.8 \\
        Unmasked AffEx & 3.8 / 7.0 & 1.4 / 3.2 & 6.2 / 9.0 & 3.6 / 7.2 \\
        LIME & 5.2 / 7.8 & 5.8 / 8.4 & 3.6 / 6.4 & 5.0 / 9.8 \\
        Gaussian Noise Mask & 7.0 / 9.6 & 5.8 / 10.4 & 11.2 / 13.0 & 12.4 / 13.6 \\
        Random & 9.8 / 10.8 & 10.0 / 12.4 & n/a & n/a \\
        \bottomrule
    \end{tabular}
\end{table}

The knee of the best method (AffEx) lies between 1.0 and 6.2 percent across the four settings, so $p{=}0.05$ sits at it: by $p{=}0.05$ AffEx has recovered 88 to 97 percent of its plateau in every setting (70 to 91 percent at $p{=}0.01$), while Gaussian Noise Mask reaches its own 90 percent point only between 5.8 and 12.4 percent. $p{=}0.01$ sits in the steep segment, where the separation between explainers is largest: on DCAMA 5-shot, the recovered mIoU at $p{=}0.01$ is 0.58 for AffEx against 0.39 for Gaussian Noise Mask, 0.36 for LIME, and 0.20 for Random, with the same picture in the other settings. Saturation is earlier in 5-shot than in 1-shot, which has a natural reading: with five supports, the top-ranked 1 percent can be drawn from the most informative pixels across all shots.

The exact choice within these regimes does not change the comparison. Across the whole 0.2 percent grid, in every one of the four settings, one of the two AffEx variants is first at every point, and the ranking of the explainers is stable up to three kinds of events: flicker at the very first grid point, swaps between the two AffEx variants where they differ by less than 0.005 (on DMTNet the two variants are equivalent, cf.\ \cref{tab:interpretability_miou}), and a single crossing between two baselines (LIME and Gaussian Noise Mask on DCAMA 5-shot around $p{=}0.042$, with the two curves within 0.02). This is the same tail-oscillation effect that motivates the two-point evaluation over full-curve areas: once curves have flattened, their ordering is determined by noise, no longer by attribution quality.

\section{Ablation Study on Support Mask Processing}
\label{sec:mask_processing}

AffEx assigns attribution signs based on the support mask: foreground pixels receive positive contributions, background pixels negative ones. To avoid abrupt sign transitions at mask boundaries and to account for the spatial diffusion introduced by convolutional and attention-based encoders, whose feature responses naturally extend slightly beyond object boundaries, we apply Gaussian smoothing to the support mask before sign assignment. Without smoothing, pixels near the object boundary may be incorrectly assigned a negative sign despite contributing positively to the prediction.

\begin{figure}[!tb]
    \centering
    \includegraphics[width=\linewidth]{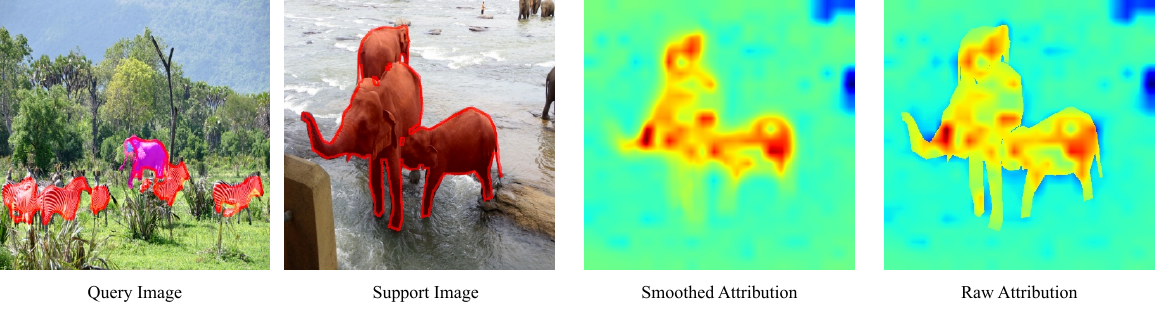}
    \caption{Qualitative comparison between smoothed and raw attribution maps. The raw attribution exhibits abrupt transitions at mask boundaries, causing pixels near the object boundary to receive erroneously negative attribution (blue). The smoothed attribution produces more gradual transitions, better reflecting the spatial uncertainty of encoder representations.}
    \label{fig:smoothing}
\end{figure}

To prevent small objects from being suppressed by the smoothing operation, we first dilate the mask before applying the Gaussian filter. Dilation uses a $7{\times}7$ square kernel applied for 5 iterations, each of which expands the foreground region by three pixels in all directions. This ensures that thin structures and small regions retain their foreground attribution even after smoothing reduces their effective extent. \Cref{fig:smoothing} illustrates the qualitative effect of mask smoothing: the raw attribution exhibits abrupt transitions at mask boundaries, causing pixels just outside the object to receive erroneously negative attribution, whereas the smoothed attribution produces more gradual and visually coherent transitions.

\Cref{tab:ablation_smoothing} reports mIoULoss@0.01 on Pascal $5^{i}$ using DCAMA in a 1-way 5-shot setting under three configurations: no processing (binary mask), Gaussian smoothing only, and dilation followed by Gaussian smoothing (the configuration used in the main experiments). The results show that Gaussian smoothing alone degrades performance relative to the binary mask baseline (13.90 vs.\ 12.01), as it suppresses the attribution of small objects by reducing their effective foreground extent. Dilation compensates for this effect by expanding the foreground region before smoothing, recovering and slightly improving over the binary mask baseline (11.56). The quantitative differences between configurations are small; the main reason for this design choice is qualitative: smoother attribution boundaries better reflect the spatial uncertainty inherent in encoder representations and produce more visually coherent attribution maps.

\begin{table}[!tb]
    \centering
    \caption{Ablation study on support mask processing. mIoULoss@0.01 on Pascal $5^{i}$ with DCAMA in a 1-way 5-shot setting.}
    \label{tab:ablation_smoothing}
    \begin{tabular}{cc|c}
        \toprule
        Dilation & Smoothing & mIoULoss@0.01 \\
        \midrule
        & & 12.01 \\
        & \checkmark & 13.90 \\
        \checkmark & \checkmark & \textbf{11.56} \\
        \bottomrule
    \end{tabular}
\end{table}

\section{An Example of Model Diagnosis with AffEx}
\label{sec:diagnosis}

\begin{figure}[!tb]
    \centering
    \includegraphics[width=\linewidth]{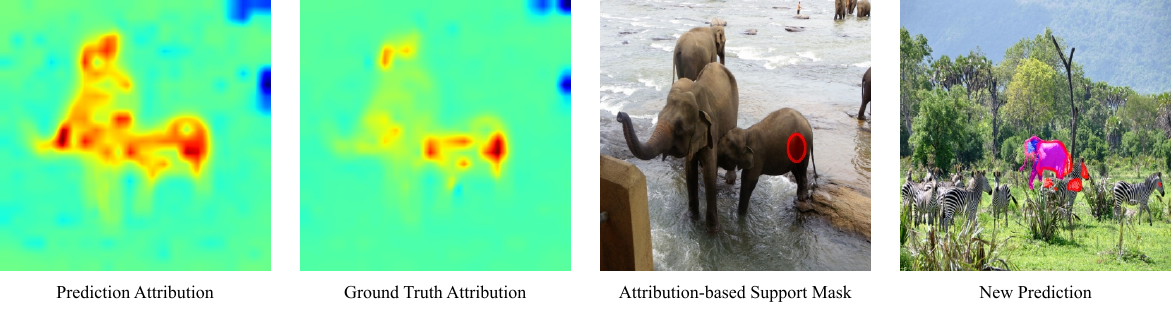}
    \caption{Model diagnosis with AffEx on a failure case of DCAMA on COCO $20^{i}$. From left to right: attribution map computed over the predicted region, attribution map computed over the ground truth region, attribution-based support mask highlighting the most influential area (circled in red), and the resulting prediction after support mask refinement. Restricting the support mask to the most influential region significantly reduces false positives on the zebras while preserving the elephant segmentation.}
    \label{fig:failure}
\end{figure}

To illustrate how AffEx can be used for model diagnosis, we present a case study on a failure mode of DCAMA \cite{shiDenseCrossQueryandSupportAttention2022} on the COCO $20^{i}$ dataset, using the same episode shown in \cref{fig:smoothing}. In this episode, the model correctly identifies the elephant in the query image but also incorrectly segments several zebras. To understand this failure, we compare the attribution maps computed over the predicted region and over the ground truth region, as shown in \cref{fig:failure}. In the former, the most influential areas are the face and the back of the elephant; in the latter, only the back is highlighted. The attribution map thus directs our attention to the query image, where we observe that the elephant's face is occluded: the only visible region that correlates with the ground truth is the back. Including the face in the support mask therefore confuses the model, as it introduces a feature that cannot be matched in the query, ultimately leading to increased false positives on the zebras. By restricting the support mask to the back region alone, the zebra false positives are partially reduced while the elephant segmentation is preserved.

The original support mask is noisy, as it does not include the background elephants. To verify that this omission does not affect the failure mode, we tested a refined support mask that explicitly includes them. The resulting prediction is very similar to the one shown in \cref{fig:smoothing}, confirming that the dominant factor is the inclusion of the face region, which the model cannot match in the query image due to occlusion.

This case study illustrates how AffEx can guide model diagnosis: by comparing attribution maps under different regions of interest, it reveals which support features drive incorrect predictions. This information can be exploited at different levels, from targeted refinement of the support mask at inference time to identifying systematic biases that may inform architectural improvements or training strategies.

\section{Decoder-Dominated Pipelines}
\label{sec:decoder_dominated}
\label{sec:vfm_models}

Recent FSS pipelines built on vision foundation models all contain an explicit dense matching stage that fits the Encoder--Match--Seg template of \cref{sec:methodology} and that AffEx can read directly; INSID3 and SANSA are two instances evaluated in \cref{tab:interpretability_miou}. The pipelines differ in what follows the matching. In GF-SAM \citep{zhangBridgePointsGraphbased2024} and Matcher \citep{liu2024matchersegmentshotusing}, the matching does not produce the mask: it seeds point prompts for a promptable SAM decoder, which then segments whole objects on its own. The final mask is produced by a decoder that consumes the entire prompted region, and the matching contributes evidence rather than the decision. Gradient-based attribution is inapplicable, as the decision paths of both pipelines contain discrete steps (point-graph clustering in GF-SAM, Hungarian matching in Matcher).

\Cref{tab:decoder_dominated} shows the consequence. The Gaussian Noise Mask baseline wins in seven of the eight settings at @0.01, the exception being Matcher on COCO $20^{i}$ 5-shot: perturbing the whole annotated region is the most damaging intervention, because the decoder relies on the region as a whole rather than on a discriminative subset of pixels. This marks the boundary of what a matching read-out can explain when a decoder reshapes the output. Read through the causal metric, the same result is a diagnostic. When AffEx scores well, as on the four models of \cref{tab:interpretability_miou}, the decision is carried by the matching; when it cedes to the mask baseline, the decision has moved into the decoder. Even there, AffEx remains the best model-dependent attribution in \cref{tab:decoder_dominated}. Covering such pipelines end to end would require pairing it with an explainer for the decoder, which we leave to future work.

\paragraph{Testing the diagnosis on the similarities.}\label{sec:affinity_intervention}
The reading above makes a prediction that can be tested inside the model rather than through the attribution. If the decoder takes the decision, the prediction should not hinge on the strongest in-mask matches: as long as enough of the support mask still matches the query, the decoder segments the object. If the matching is the decision, removing those matches should degrade the prediction. We intervened where AffEx reads the model, on the support--query similarities $S$ of the decision path. Let $F$ be the in-mask support positions, with the mask taken as each model sees it on its own similarity grid, and let $k = \lceil \kappa\, n_F \rceil$, a fraction $\kappa$ of the $n_F$ in-mask positions rounded up. At every query position $j$, the $k$ highest similarities $S_{ij}$ with $i \in F$ were lowered by a fixed amount,
\begin{equation}
    \tilde S_{ij} = S_{ij} - \delta\,\sigma_S \quad \text{for } i \in \operatorname{top}_{k}\{ S_{i'j} : i' \in F \}, \qquad \tilde S_{ij} = S_{ij} \ \text{otherwise},
    \label{eq:topk_down}
\end{equation}
where $\sigma_S$ is the standard deviation of $S$. The out-of-mask similarities, the images and the support mask were left untouched, so the intervention removes the support evidence the matching ranks highest and nothing else: a deletion curve applied to the similarities instead of to the support pixels. We swept $\kappa \in \{0.01, 0.05, 0.1, 0.25, 0.5\}$ for $\delta \in \{2, 4\}$ over 1000 Pascal $5^{i}$ 1-shot episodes and report the relative mIoU loss against each model's unperturbed prediction. Every forward pass of an episode used the same seed, which also fixed Matcher's random sampling of point prompts, and all four models then reproduced their unperturbed masks exactly, so every change was caused by the intervention.

\begin{table}[!ht]
    \centering
    \small
    \caption{Similarity intervention on 1000 Pascal $5^{i}$ 1-shot episodes: relative mIoU loss (\%) when the $k$ highest in-mask similarities at every query position are lowered by $\delta\,\sigma_S$ (\cref{eq:topk_down}). Standard errors of the underlying mIoU changes are between 0.3 and 1.1 points.}
    \label{tab:affinity_intervention}
    \begin{tabular}{llrrrrrr}
        \toprule
        & Model & Clean mIoU & $\kappa{=}0.01$ & $0.05$ & $0.1$ & $0.25$ & $0.5$ \\
        \midrule
        \multirow{4}{*}{$\delta{=}2$}
          & DCAMA   & 64.5 & $-1.6$  & $-4.7$  & $-9.9$  & $-34.2$ & $-65.3$ \\
          & INSID3  & 60.7 & $-10.5$ & $-25.5$ & $-35.5$ & $-54.6$ & $-74.5$ \\
          & GF-SAM  & 61.5 & $-0.2$  & $-1.5$  & $-2.2$  & $-8.5$  & $-20.3$ \\
          & Matcher & 53.6 & $-2.4$  & $-8.0$  & $-12.7$ & $-14.6$ & $-11.7$ \\
        \midrule
        \multirow{4}{*}{$\delta{=}4$}
          & DCAMA   & 64.5 & $-1.5$  & $-4.5$  & $-9.3$  & $-31.0$ & $-57.7$ \\
          & INSID3  & 60.7 & $-10.5$ & $-25.5$ & $-35.8$ & $-56.5$ & $-82.5$ \\
          & GF-SAM  & 61.5 & $-0.2$  & $-2.9$  & $-6.9$  & $-20.8$ & $-30.6$ \\
          & Matcher & 53.6 & $-2.8$  & $-9.4$  & $-17.2$ & $-28.4$ & $-23.8$ \\
        \bottomrule
    \end{tabular}
\end{table}

\Cref{tab:affinity_intervention} separates the two regimes once the deletion is deep. At $\kappa=0.5$ and $\delta=4$, DCAMA loses $57.7\%$ of its mIoU and INSID3 $82.5\%$, and both keep falling steeply as $\kappa$ grows; GF-SAM loses $30.6\%$ and Matcher $23.8\%$. Between $\kappa=0.25$ and $\kappa=0.5$, GF-SAM loses at most 12 further points and Matcher none, against 20 to 31 for DCAMA and INSID3. With half of the strongest in-mask matches gone, the remaining ones still point the SAM decoder at the object, and the decoder segments it from there. The smaller drop gives the same picture ($65.3\%$ and $74.5\%$ against $20.3\%$ and $11.7\%$).

At small $\kappa$ the models do not sort into the two groups. INSID3 reacts at once, losing $10.5\%$ at $\kappa=0.01$, since its backward vote reads only the mask label of the best-matching support position. DCAMA, whose attention output at a query position is the softmax mass on all the in-mask keys, loses less than $10\%$ up to $\kappa=0.1$, close to GF-SAM at $\delta=4$ ($9.3\%$ against $6.9\%$) and less than Matcher ($17.2\%$). What separates matching-dominated from decoder-dominated models is therefore the shape of the curve: the former keep degrading steeply as the in-mask evidence is removed, while the latter degrade far more slowly once enough of it remains.

One caveat concerns Matcher: its loss falls between $\kappa=0.25$ and $\kappa=0.5$, and at $\kappa=0.25$ it is close to DCAMA's ($28.4\%$ against $31.0\%$ at $\delta=4$), so at deep deletion the separation rests mainly on GF-SAM, which still degrades far more slowly than the matching-dominated models. The same behavior appears from the attribution side in the causal metrics of \cref{tab:decoder_dominated}, where the Gaussian Noise Mask baseline beats every attribution that ranks individual support pixels in seven of the eight settings.

\section{AffEx on a Prototype-Based Head}
\label{sec:panet_details}
\label{sec:prototype_head}

The prototype-based head follows PANet \citep{wangPANetFewShotImage2019} on top of DCAMA's encoder: the same frozen Swin-B backbone with identical weights (DCAMA itself is trained head-only on this backbone). For each of the three deepest feature levels, support features are masked-average-pooled into a foreground and a background prototype, and each query position is classified by cosine similarity to the two prototypes, scaled by a learned temperature; the per-level logits are upsampled and averaged. The only trained parameters are per-level linear projections (initialized to the identity) and the temperature. The head was trained episodically on Pascal $5^{i}$ with the same data pipeline used for DCAMA, with cross-entropy loss and early stopping, and reached 46.6 validation mIoU in the 1-shot setting, in line with PANet. On COCO $20^{i}$, as for DCAMA, the head trained on Pascal $5^{i}$ was evaluated on the COCO episodes. The attributions were evaluated on the trained head without any change to its architecture, training, or prediction path, following the same protocol, episodes, and metrics as \cref{tab:interpretability_miou}.

Since the prototype is a weighted average of normalized support features, the foreground logit factorizes exactly as
\begin{equation}
    \cos(\tilde{q}, \hat{P}) = \frac{\sum_{s} m_s \, (\tilde{q} \cdot \tilde{v}_s)}{\big\lVert \sum_{s} m_s \, \tilde{v}_s \big\rVert},
    \label{eq:proto_factorization}
\end{equation}
with the sum over support pixels $s$, $m_s$ the mask value and $\tilde{v}_s$ the normalized feature: a mask-weighted mean of pairwise query--support similarities, divided by a constant independent of query position and support pixel. Upsampling the pixel-to-prototype similarity, as prototype-part methods do with their activation maps, gives the \emph{prototype read-out} in \cref{tab:interpretability_proto}: each support pixel is scored by the cosine similarity between its (projected, normalized) feature and the foreground prototype, per level, and the per-level maps are upsampled to image size and averaged. Reading the pairwise terms, one step before the pooling sums them, gives AffEx: the model computes the sum, the explanation looks at the summands. The two are the same quantity at two levels of aggregation: the read-out is query-blind by construction, being what remains of the pairwise matrix once the query axis is replaced by the support average.

Three observations follow from \cref{tab:interpretability_proto}. First, AffEx or its unmasked variant is the best attribution at the early point in every setting, with the prototype read-out close behind but consistently worse; the gap isolates the value of conditioning on the query, which the read-out never sees. Second, AffEx behaves consistently across heads sharing the same encoder (21.84/9.95 on the prototype head vs.\ 20.22/6.79 on DCAMA's attention head at 1-shot on Pascal $5^{i}$), supporting its reading as a general read-out of the matching stage, with the prototype case as the special case where matching is against a pooled vector. Third, one exception: at 5-shot with $p{=}0.05$, LIME comes out best on both datasets; the effect is specific to the prototype head, does not appear on the matching architectures, and at the early point the ordering is unambiguous on both heads.

\section{Additional Qualitative Samples}
\label{sec:qualitative_extra}

\begin{figure*}[!tb]
    \centering
    \includegraphics[width=\linewidth]{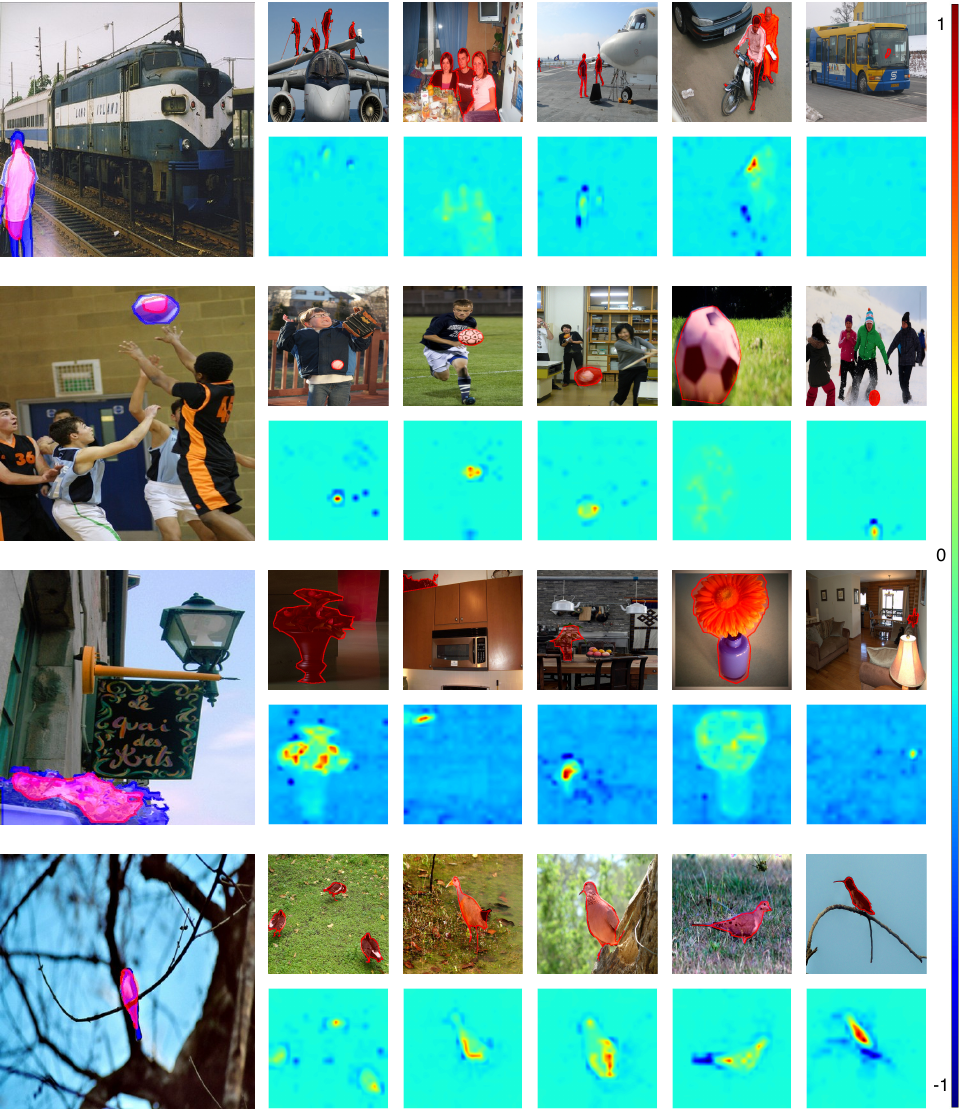}
    \caption{
    Qualitative results for AffEx with DCAMA across multiple episodes from both COCO $20^{i}$ and Pascal $5^{i}$. For each episode, the left column shows the query image with the predicted segmentation in red and the ground truth overlaid in blue. The right column presents the corresponding support set alongside the attribution maps produced by the proposed method.
    }
    \label{fig:qualit_dcama}
\end{figure*}

\begin{figure*}[!tb]
    \centering
    \includegraphics[width=\linewidth]{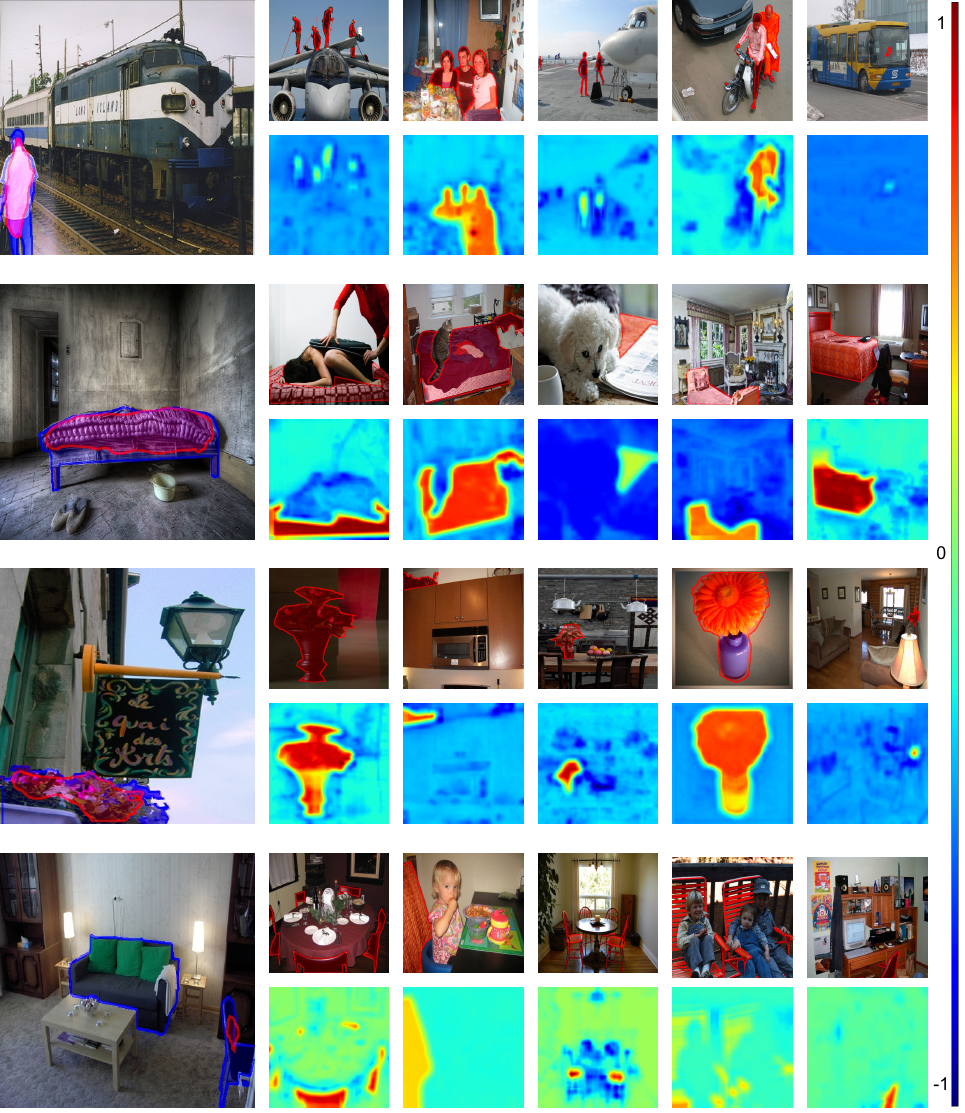}
    \caption{
    Qualitative results for AffEx with DMTNet across multiple episodes from both COCO $20^{i}$ and Pascal $5^{i}$. For each episode, the left column shows the query image with the predicted segmentation in red and the ground truth overlaid in blue. The right column presents the corresponding support set alongside the attribution maps produced by the proposed method.
    }
    \label{fig:qualit_dmtnet}
\end{figure*}

\begin{figure*}[!tb]
    \centering
    \includegraphics[width=\linewidth]{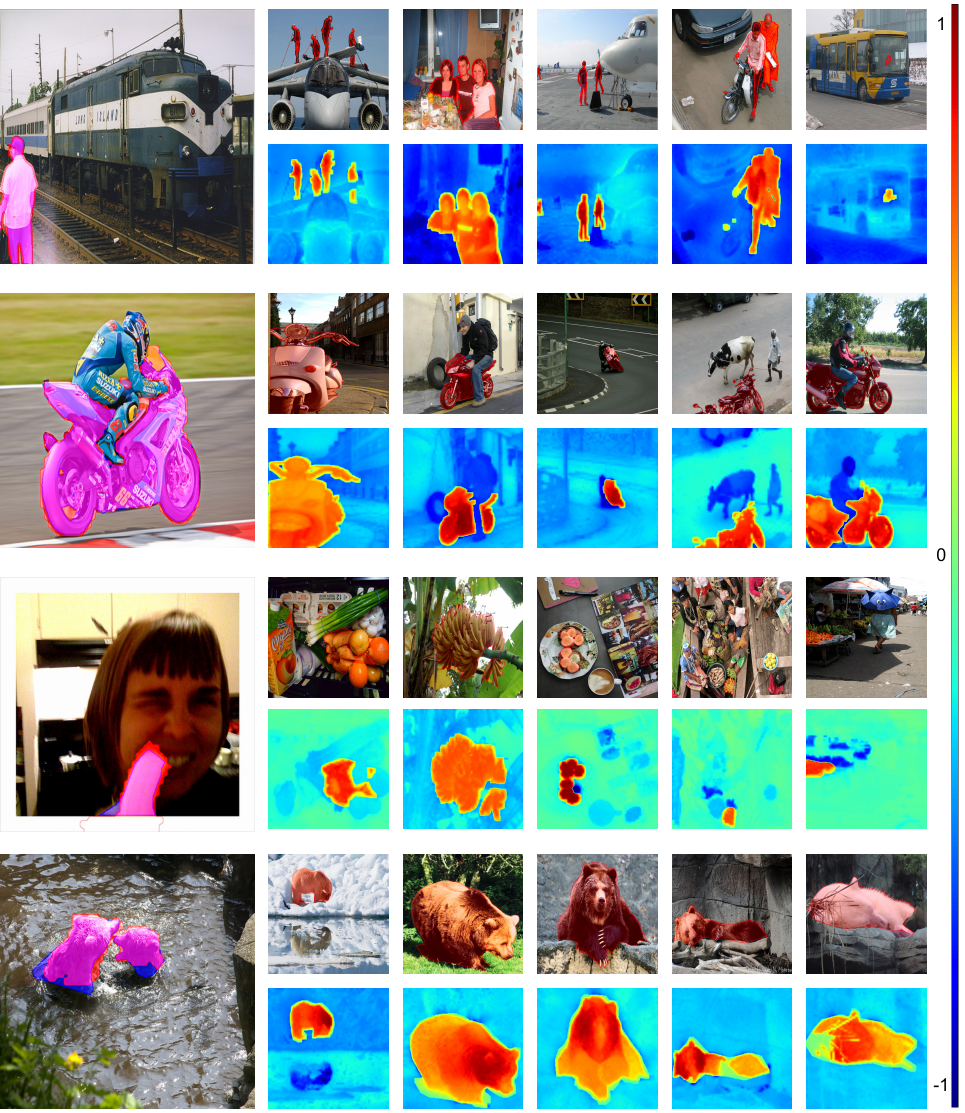}
    \caption{
    Qualitative results for AffEx with INSID3 across multiple episodes from both COCO $20^{i}$ and Pascal $5^{i}$. For each episode, the left column shows the query image with the predicted segmentation in red and the ground truth overlaid in blue. The right column presents the corresponding support set alongside the attribution maps produced by the proposed method. The first episode is shared with \cref{fig:qualit_dcama}, \cref{fig:qualit_dmtnet} and \cref{fig:qualit_sansa}.
    }
    \label{fig:qualit_insid3}
\end{figure*}

\begin{figure*}[!tb]
    \centering
    \includegraphics[width=\linewidth]{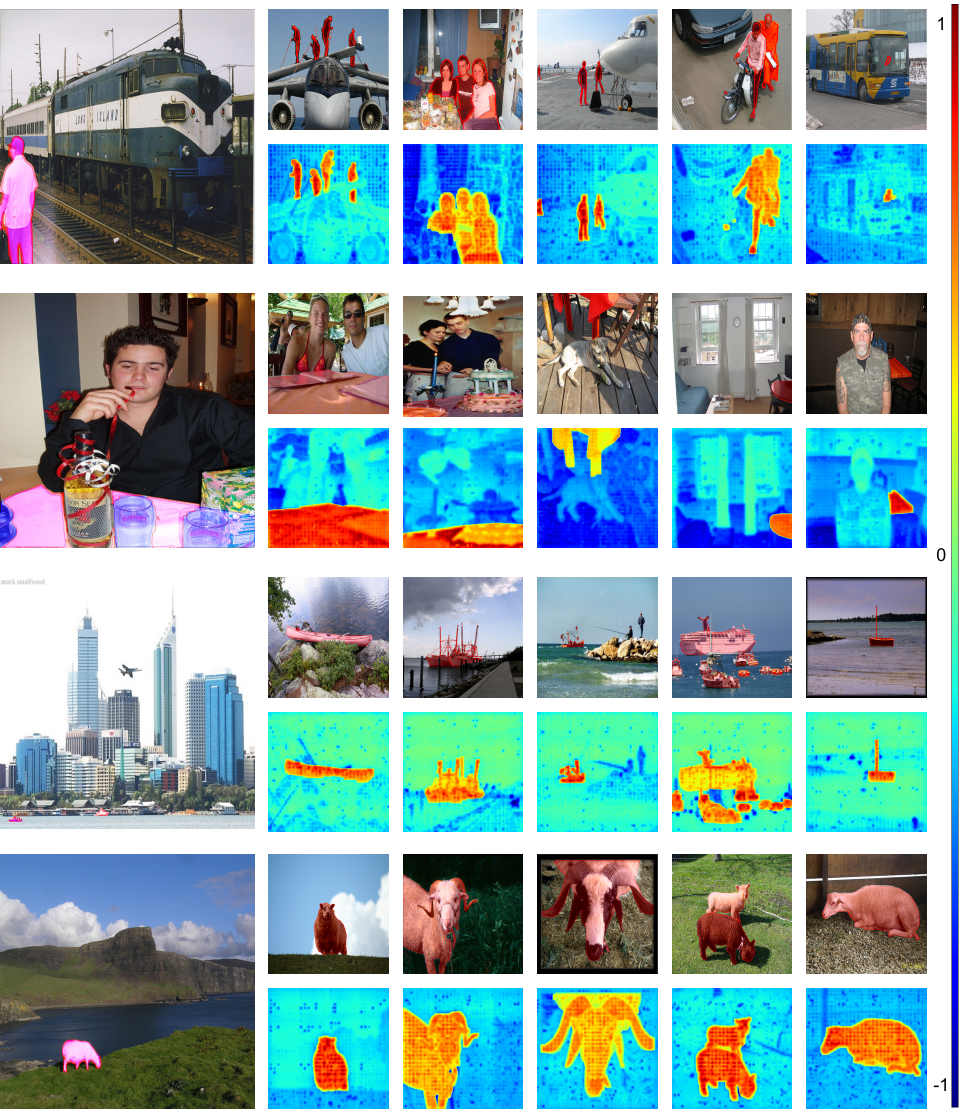}
    \caption{
    Qualitative results for AffEx with SANSA across multiple episodes from both COCO $20^{i}$ and Pascal $5^{i}$. For each episode, the left column shows the query image with the predicted segmentation in red and the ground truth overlaid in blue. The right column presents the corresponding support set alongside the attribution maps produced by the proposed method. The first episode is shared with \cref{fig:qualit_dcama}, \cref{fig:qualit_dmtnet} and \cref{fig:qualit_insid3}.
    }
    \label{fig:qualit_sansa}
\end{figure*}

\begin{figure*}[!tb]
    \centering
    \includegraphics[width=\linewidth]{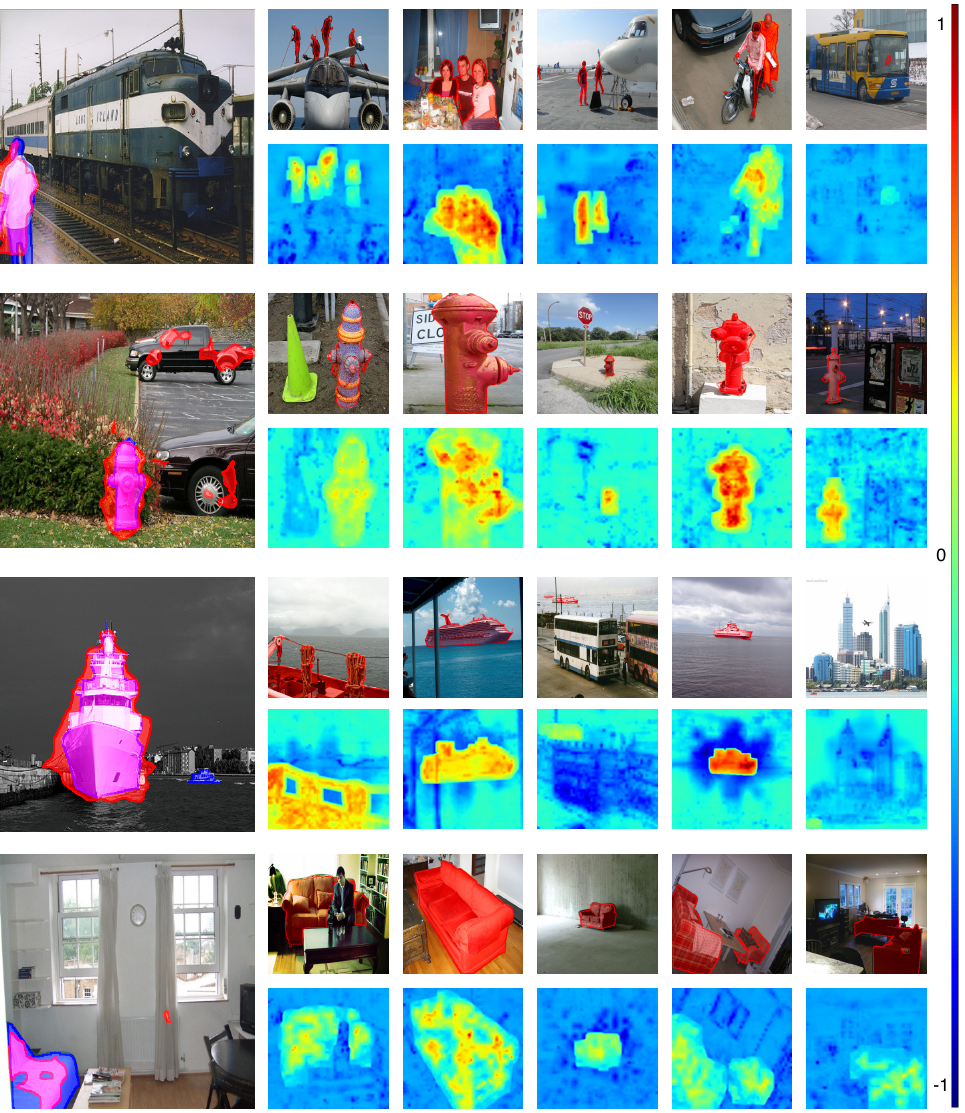}
    \caption{
    Qualitative results for AffEx with the prototype head of \cref{sec:prototype_head} across multiple episodes from both COCO $20^{i}$ and Pascal $5^{i}$. For each episode, the left column shows the query image with the predicted segmentation in red and the ground truth overlaid in blue. The right column presents the corresponding support set alongside the attribution maps produced by the proposed method. The first episode is shared with \cref{fig:qualit_dcama}, \cref{fig:qualit_dmtnet}, \cref{fig:qualit_insid3} and \cref{fig:qualit_sansa}.
    }
    \label{fig:qualit_panet}
\end{figure*}

This section presents additional qualitative examples complementing the analysis of the main paper. All visualizations were generated using the AffEx method. \Cref{fig:qualit_dcama} shows examples produced with DCAMA, \cref{fig:qualit_dmtnet} reports the corresponding results for DMTNet, \cref{fig:qualit_insid3} and \cref{fig:qualit_sansa} those for INSID3 and SANSA, and \cref{fig:qualit_panet} those for the prototype head of \cref{sec:prototype_head}, evaluated on the COCO $20^{i}$ \cite{nguyen2019feature} and Pascal $5^{i}$ \cite{shabanOneShotLearningSemantic2017} datasets. Each figure includes a series of query--support pairs illustrating the models' predicted segmentations together with their corresponding attribution maps. The first and third episodes are shared between \cref{fig:qualit_dcama} and \cref{fig:qualit_dmtnet} to facilitate a direct comparison of how attributions differ across models. The first episode is common to all five figures, so that the same query and the same five support images can be read across every model.

With five supports, the maps answer two questions. Within a support image they show which parts of the annotated object the prediction rested on, and since the maps of an episode share a color scale, they also show which shot carried it: a shot that contributed little stays flat next to one that drives the segmentation. They further expose evidence gathered outside the annotated region. A response in the background of a support image is a match the model found on context rather than on the object, which the mask rule of \cref{sec:interpretation} signs as negative, so a practitioner can tell an interfering background from a useful one and revise the support set accordingly.

Read across the figures, the same episode turns the maps into a comparison of the matching mechanisms themselves, in what each model responds to and in how sharply it localizes it. The attributions of INSID3, whose correspondence is computed on DINOv3 features, follow object contours closely and separate individual instances, including small and partially occluded ones. SANSA, matching dense SAM2 features, is comparably sharp and covers whole objects, with a regular grid-like texture across the whole map. The ResNet-based correlation of DMTNet produces visibly coarser, blob-like maps that spread over an object and part of its surroundings, while the attention scores of DCAMA concentrate on a few small regions rather than covering the object. The progression is visible on the shared episode alone, and it is consistent with the generational gap between the encoders on which these models are built.

\section{IAUC and DAUC Plots}
\label{sec:auc_examples}
\begin{figure*}[!tb]
    \centering
    \includegraphics[width=\linewidth]{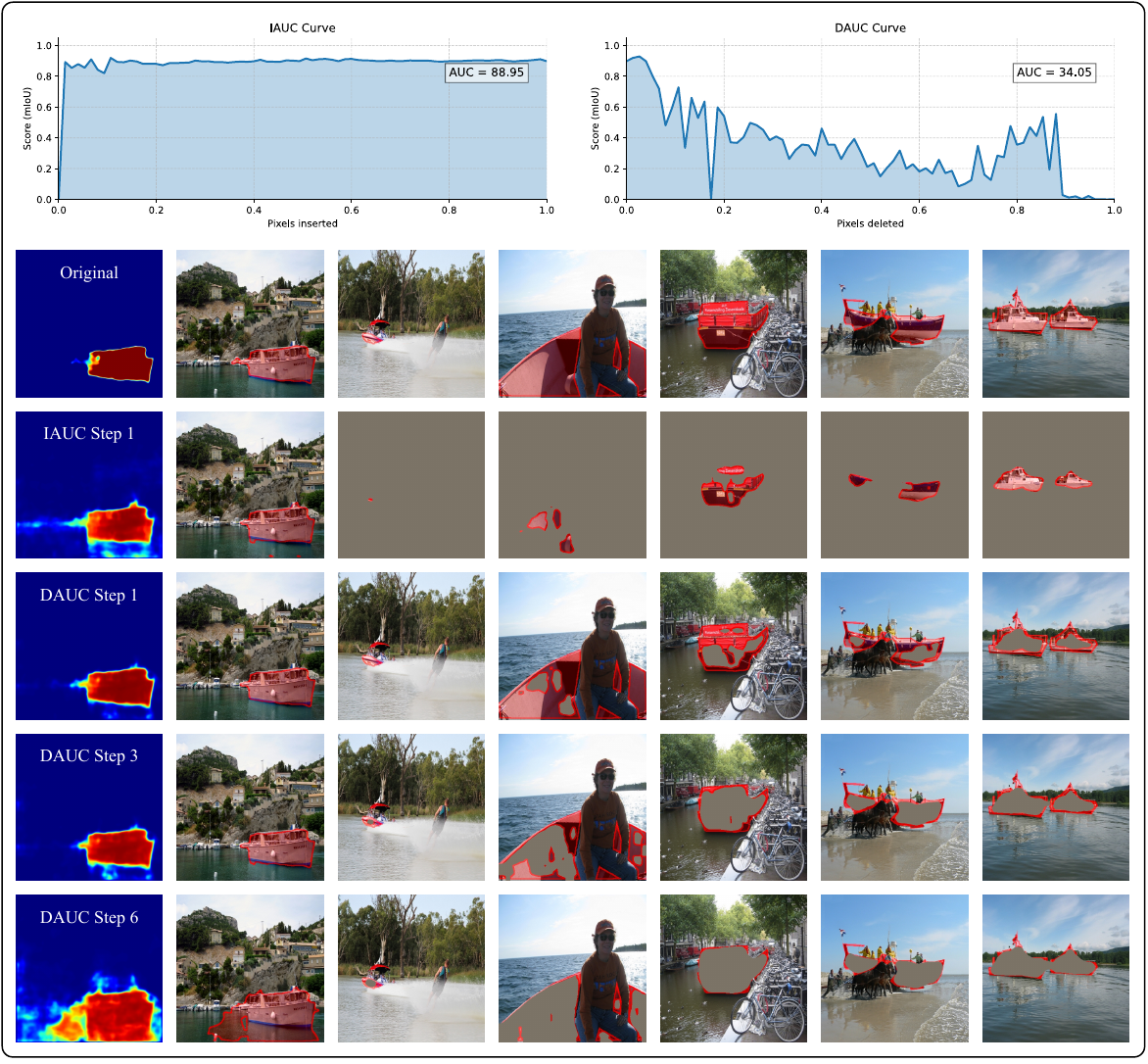}
    \caption{
    A visually clean episode of the insertion and deletion evaluation for DCAMA on COCO $20^{i}$ (5-shot) using AffEx. 
    The top row reports the IAUC and DAUC curves for a case with a clearly delineated foreground object. 
    Subsequent rows show the confidence map, the predicted query mask (red) overlaid with the ground truth (blue), and the support images with their masks at different insertion/deletion steps, illustrating stable attribution behavior.
    }
    \label{fig:iauc_dcama1}
\end{figure*}

\begin{figure*}[!tb]
    \centering
    \includegraphics[width=\linewidth]{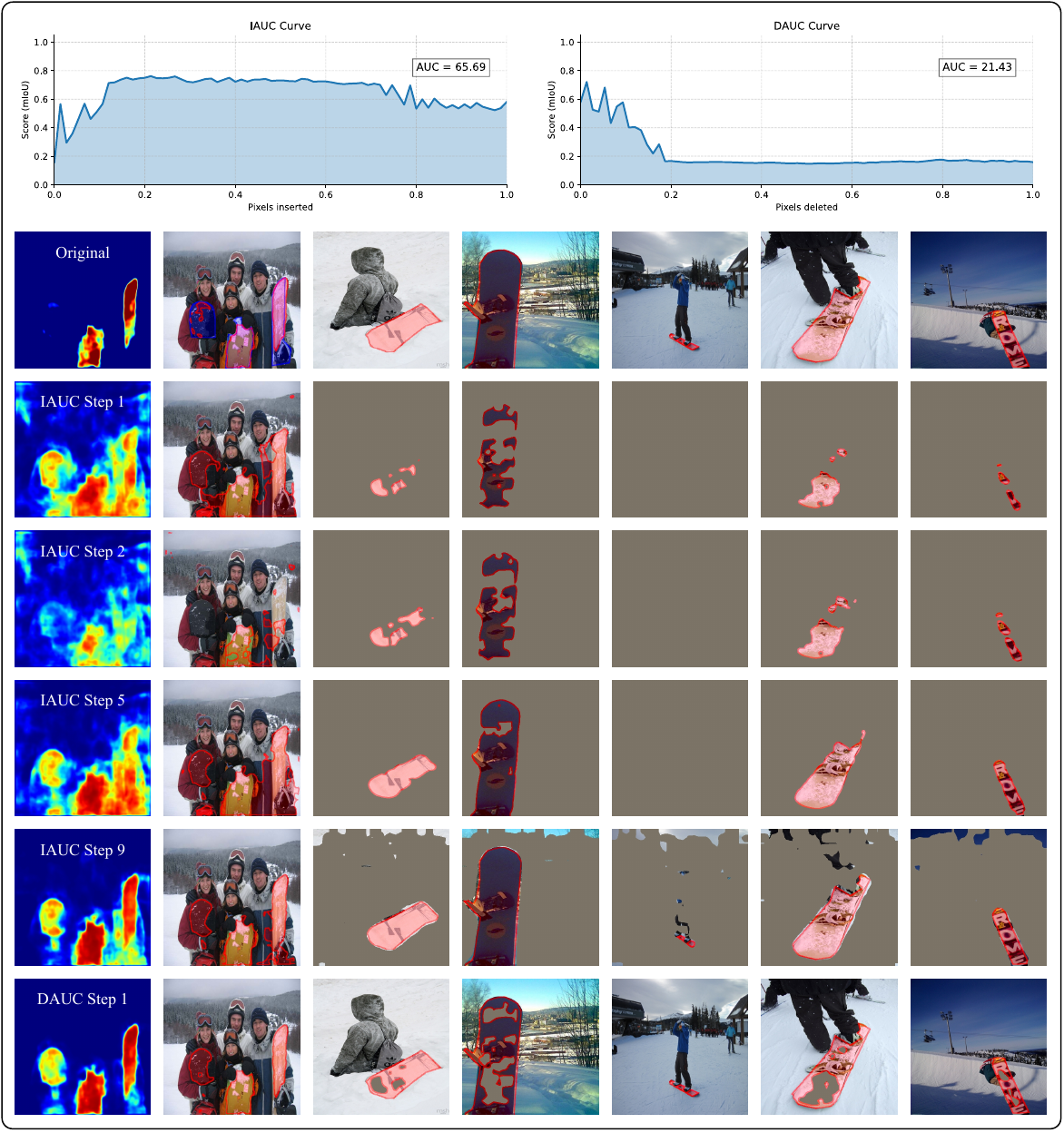}
    \caption{
    A challenging episode highlighting unstable insertion and deletion behavior for DMTNet on COCO $20^{i}$ (5-shot) with AffEx. 
    The top row presents IAUC and DAUC curves reflecting the increased difficulty of this scenario. 
    The lower rows display the confidence map, the predicted query mask (red) against the ground truth (blue), and the support images with their masks across insertion/deletion steps, revealing fluctuations in evidence attribution.
    }
    \label{fig:iauc_dmtnet}
\end{figure*}

\begin{figure*}[!tb]
    \centering
    \includegraphics[width=\linewidth]{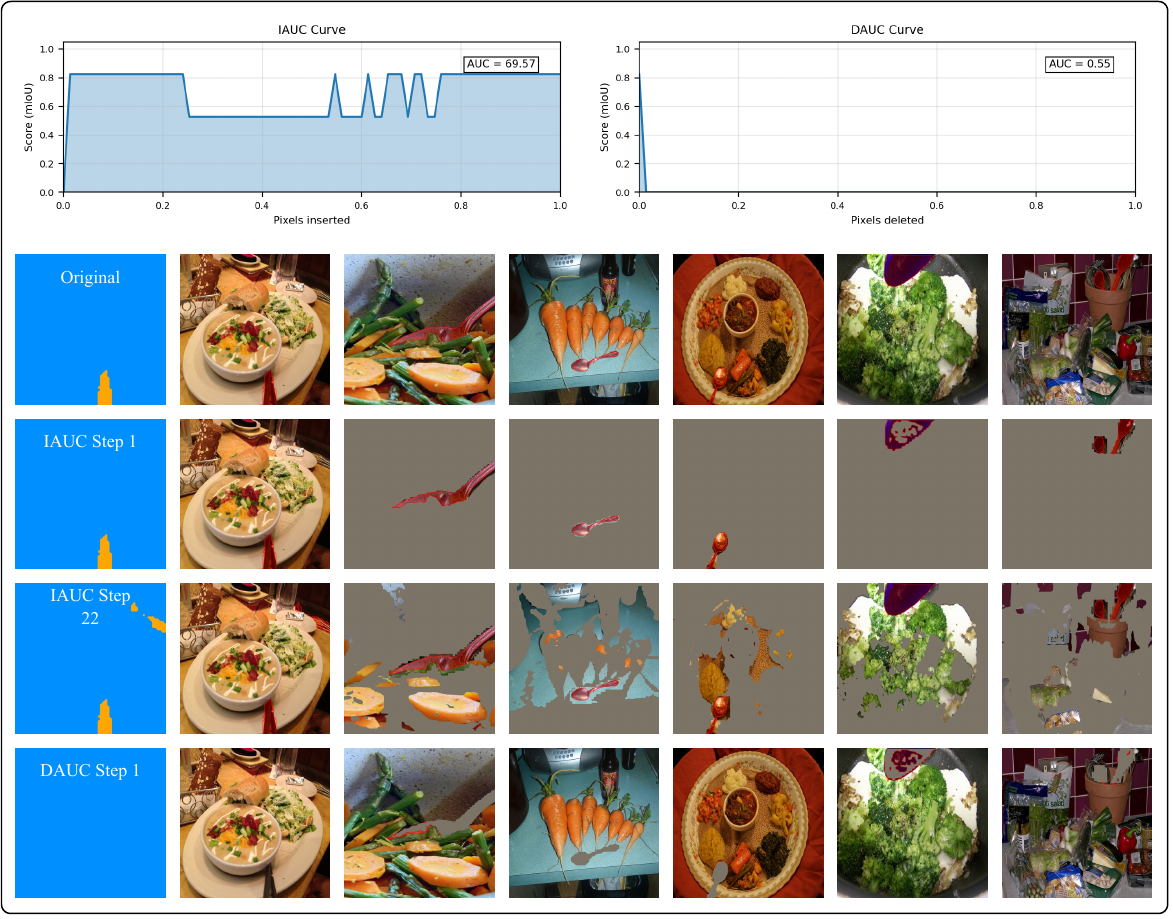}
    \caption{An episode of the insertion and deletion evaluation for INSID3 on COCO $20^{i}$ (5-shot) using AffEx. The top row reports the IAUC and DAUC curves. Subsequent rows show the model output, the predicted query mask (red), and the support images with their masks at different insertion/deletion steps. INSID3 returns a hard mask, so the first column is its binary output rather than a confidence map.}
    \label{fig:iauc_insid3}
\end{figure*}

\begin{figure*}[!tb]
    \centering
    \includegraphics[width=\linewidth]{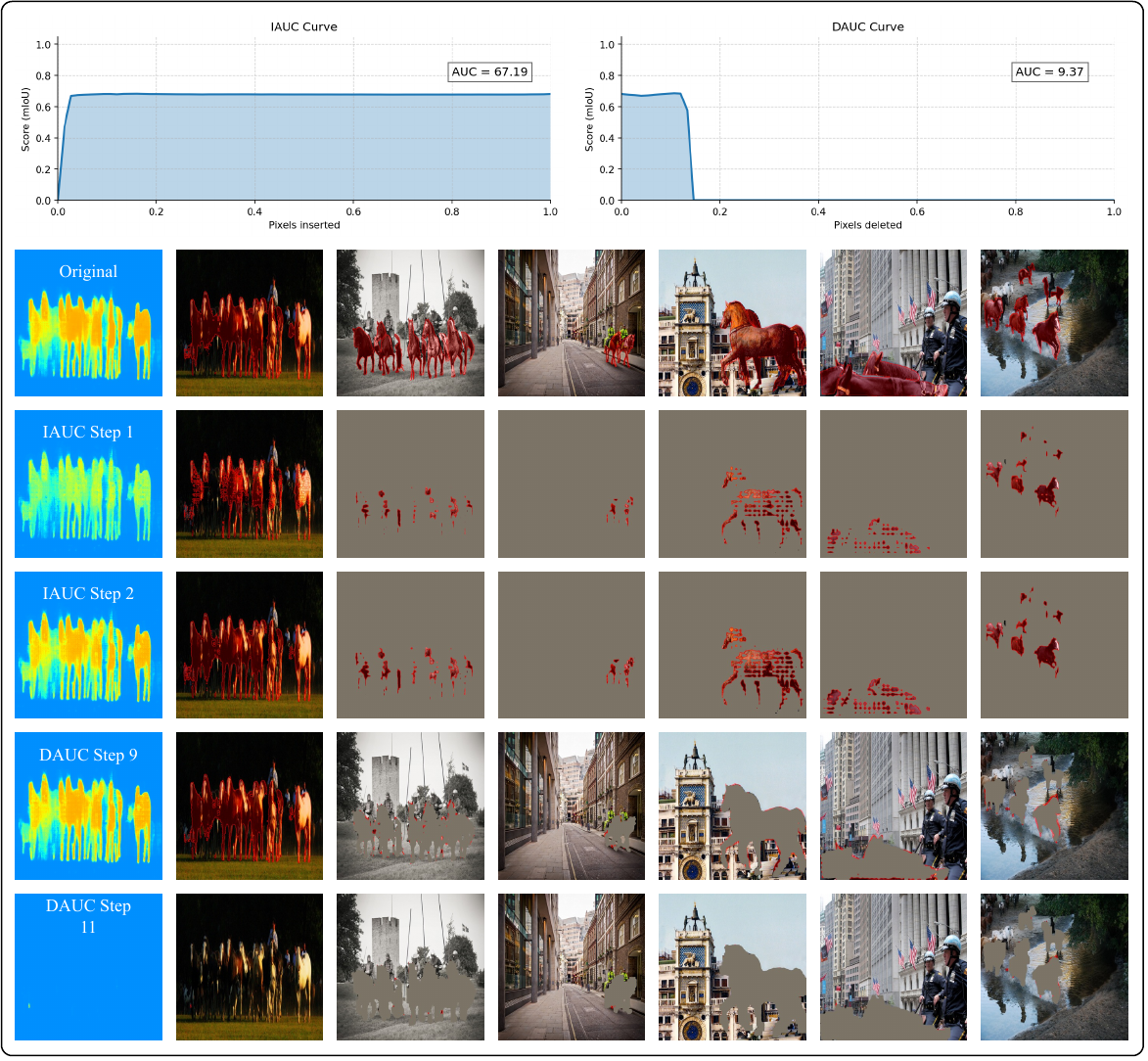}
    \caption{An episode of the insertion and deletion evaluation for SANSA on COCO $20^{i}$ (5-shot) using AffEx. The top row reports the IAUC and DAUC curves. Subsequent rows show the confidence map, the predicted query mask (red), and the support images with their masks at different insertion/deletion steps.}
    \label{fig:iauc_sansa}
\end{figure*}

\begin{figure*}[!tb]
    \centering
    \includegraphics[width=\linewidth]{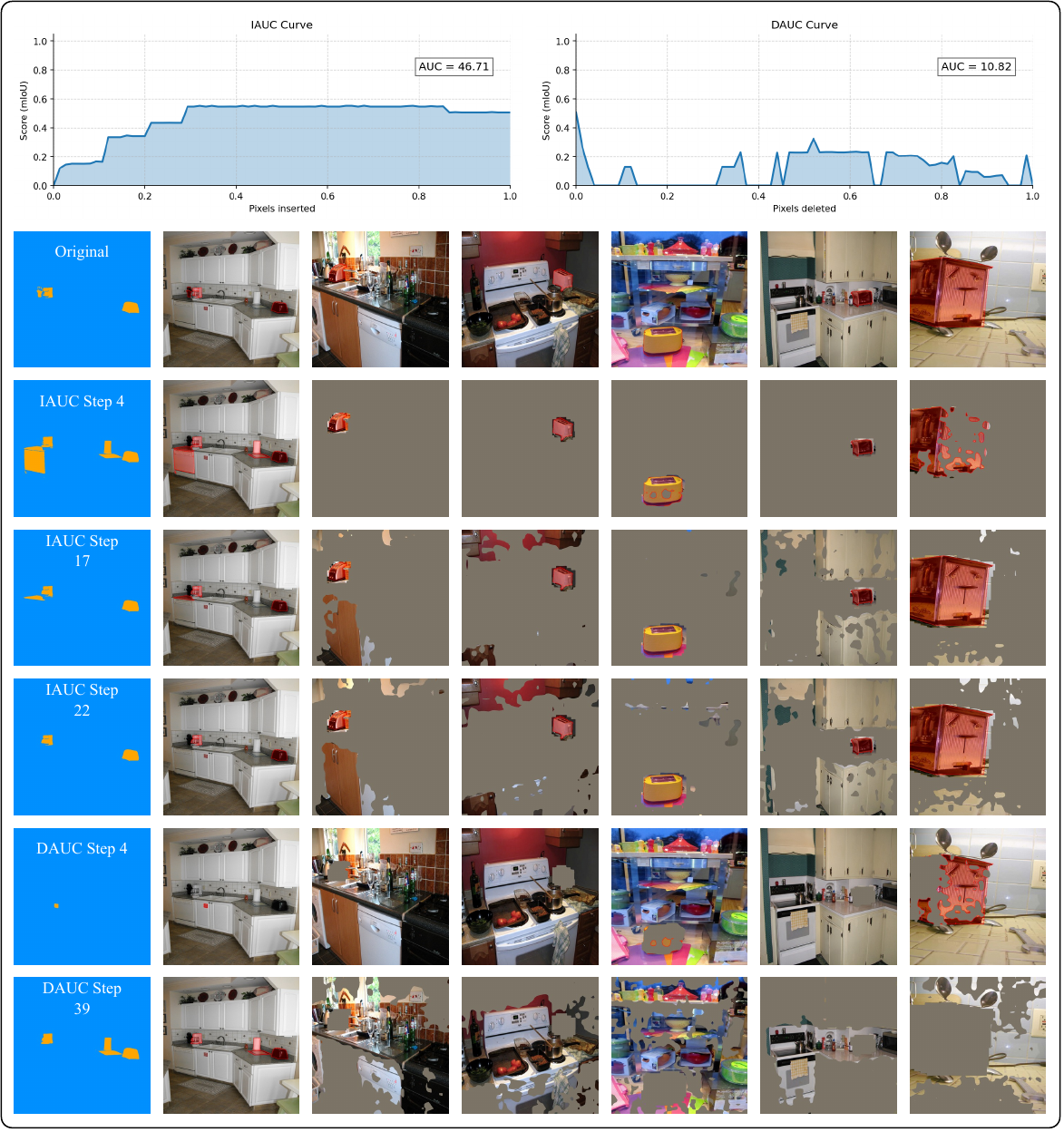}
    \caption{An episode of the insertion and deletion evaluation for GF-SAM on COCO $20^{i}$ (5-shot) using AffEx. The top row reports the IAUC and DAUC curves. Subsequent rows show the model output, the predicted query mask (red), and the support images with their masks at different insertion/deletion steps. GF-SAM returns a hard mask, so the first column is its binary output rather than a confidence map.}
    \label{fig:iauc_gfsam}
\end{figure*}

\begin{figure*}[!tb]
    \centering
    \includegraphics[width=\linewidth]{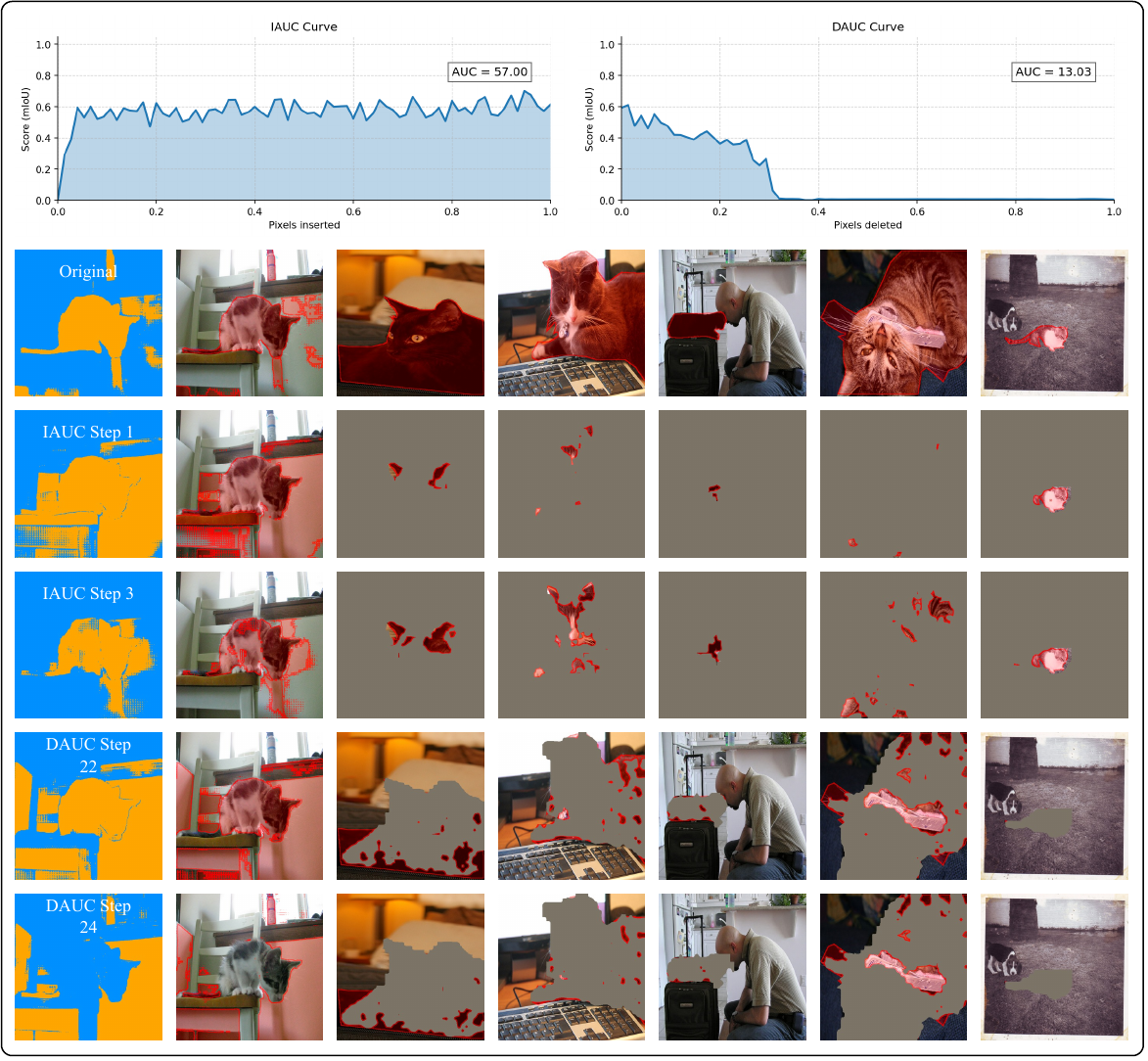}
    \caption{An episode of the insertion and deletion evaluation for Matcher on COCO $20^{i}$ (5-shot) using AffEx. The top row reports the IAUC and DAUC curves. Subsequent rows show the model output, the predicted query mask (red), and the support images with their masks at different insertion/deletion steps. Matcher returns a hard mask, so the first column is its binary output rather than a confidence map.}
    \label{fig:iauc_matcher}
\end{figure*}

\begin{figure*}[!tb]
    \centering
    \includegraphics[width=\linewidth]{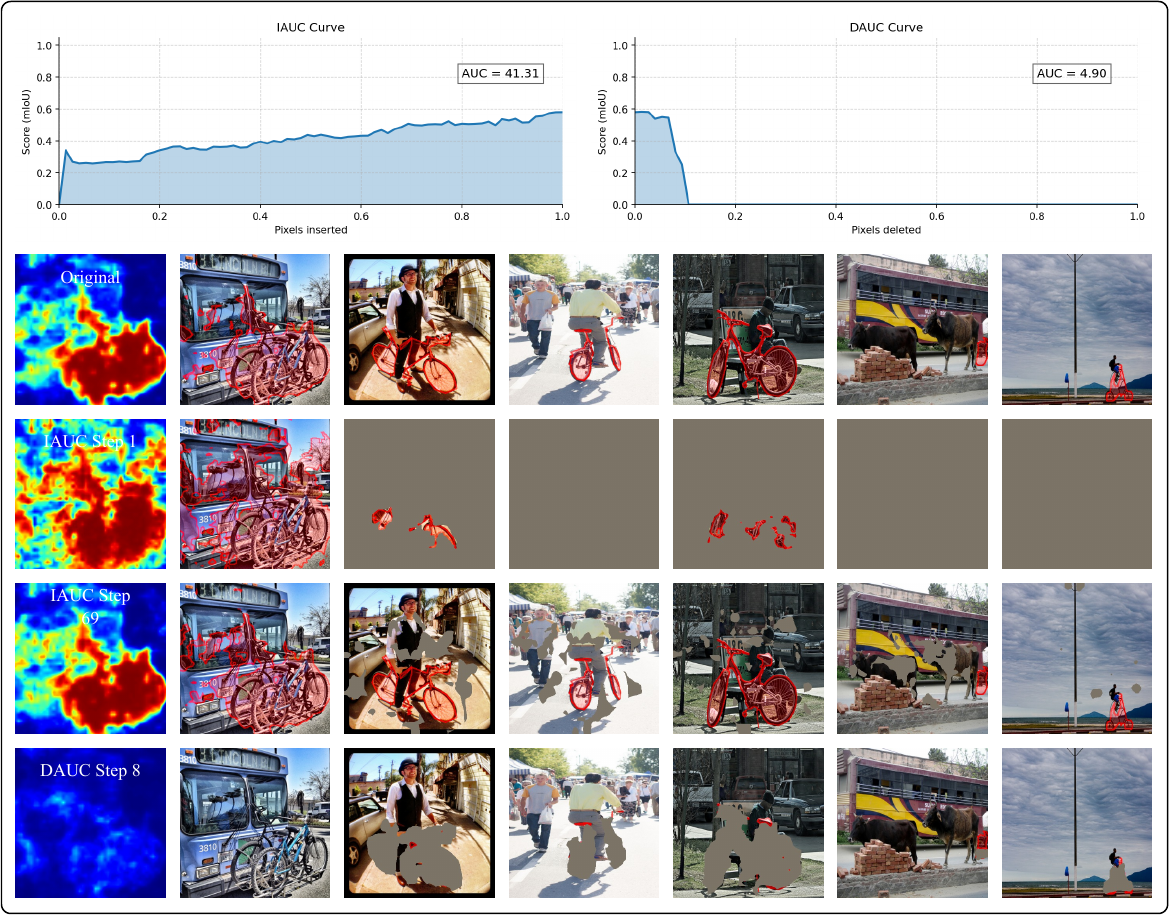}
    \caption{An episode of the insertion and deletion evaluation for the prototype head of \cref{sec:prototype_head} on COCO $20^{i}$ (5-shot) using AffEx. The head is the one trained on Pascal $5^{i}$ with fold 0 held out, so bicycle is a class it never saw in training. The top row reports the IAUC and DAUC curves. Subsequent rows show the confidence map, the predicted query mask (red), and the support images with their masks at different insertion/deletion steps.}
    \label{fig:iauc_panet}
\end{figure*}

To better illustrate how IAUC and DAUC reflect attribution quality, we provide additional visualizations of their corresponding curves. \Cref{fig:iauc_dcama1,fig:iauc_dmtnet} show one example each of Insertion and Deletion curves for DCAMA and DMTNet on COCO $20^{i}$, and \cref{fig:iauc_insid3,fig:iauc_sansa,fig:iauc_gfsam,fig:iauc_matcher,fig:iauc_panet} one episode each for INSID3, SANSA, GF-SAM, Matcher and the prototype head of \cref{sec:prototype_head}. Each example includes the IAUC and DAUC plots, the model's confidence map, the segmented query image (prediction in red, ground truth in blue), and the support set with corresponding masks (in red). For the first episode in \cref{fig:iauc_dcama1}, we display only the initial IAUC step, as the remainder of the curve is nearly flat. In this case, the pixels selected by AffEx quickly lead to a high-quality segmentation, producing a steep IAUC curve. The DAUC curve, by contrast, decreases more gradually and exhibits small oscillations as pixels are removed, indicating that the model can still rely on the remaining support information to maintain a reasonable prediction. This is further confirmed by the DAUC-step confidence maps: even after removing the most influential pixels, the model retains partial segmentation ability, suggesting either an inherent foreground bias or a dependence on broader contextual cues (e.g., the sea) that guide the prediction toward the boat. This ability fluctuates across the DAUC steps, but the model never completely loses its ability to segment the object until 90\% of the pixels are removed.
Such episodes are common in the 5-shot setting, which motivates the use of IAUC@p and mIoULoss@p to better capture early differences in the curves.

The second episode, on DMTNet, in \cref{fig:iauc_dmtnet} shows a slower IAUC rise, accompanied by fluctuations in the confidence maps. As this episode is more challenging, the model requires a larger fraction of inserted pixels to achieve a satisfactory segmentation, leading to a less pronounced IAUC slope. Conversely, removing only 20\% of the most important pixels already reduces the score below 0.2, producing a steep DAUC curve. Unlike the first episode, the confidence maps in the early DAUC steps immediately deteriorate, and the model begins to highlight and segment different, inconsistent regions.

On the INSID3 episode of \cref{fig:iauc_insid3}, inserting the $1.3\%$ of support pixels that AffEx ranks highest restores the full prediction (mIoU $0.82$), and deleting the same pixels removes it, so the DAUC curve drops to zero at the first step. The IAUC curve takes only three values: INSID3 decides through a discrete vote followed by clustering, so as pixels are inserted its prediction switches between two segmentations (mIoU $0.53$ and $0.82$) rather than changing gradually. The drop to $0.53$, which starts at step 19 and is visible at step 22 in the figure, comes from a fork in the background that the prediction picks up as a false positive. Since INSID3 returns a hard mask, the first column of the figure shows its binary output instead of a confidence map. On the SANSA episode of \cref{fig:iauc_sansa}, the prediction reaches $90\%$ of its plateau after $2.7\%$ of the pixels are inserted and stays flat afterwards. Under deletion it holds while the first $12\%$ of the ranked pixels are removed, with confidence maps almost unchanged, and then fails abruptly: removing the next $2.7\%$ of the support pixels, from step 9 to step 11, erases the whole segmentation and takes the mIoU from $0.68$ to zero. Instead of shrinking as the evidence is removed, the prediction stays intact until too little is left, and then disappears at once.

On the GF-SAM episode of \cref{fig:iauc_gfsam}, the insertion curve rises in stairs: the mIoU stays near $0.15$ up to step 8, then jumps to $0.33$ at step 9, to $0.43$ at step 16 and to $0.55$ at step 22, and barely moves in between. Inserted pixels change the prediction only when they change the point prompts the SAM decoder receives, and a new prompt adds a whole region at once. Deletion removes the prediction almost immediately, from $0.51$ to zero by step 3 ($4\%$ of the pixels). The curve then oscillates between zero and $0.24$, with a spike to $0.33$ at step 39, when about half of the support pixels are gone: from the evidence that is left, the matching can still place a prompt on the object, and the decoder grows it into a region.

On the Matcher episode of \cref{fig:iauc_matcher}, inserting the $1.3\%$ of support pixels that AffEx ranks highest, small fragments of the cats (mostly eyes and ears), is already enough for Matcher to find the query cat. The prediction also covers the chair and the table (mIoU $0.29$). At $4\%$ (step 3) the mIoU reaches $0.59$, the value obtained with the full support set, and the curve stays on that plateau from then on. Part of its oscillation comes from Matcher's random sampling of point prompts: the full support set, seen at DAUC step 0 and at IAUC step 75, gives $0.59$ and $0.61$. Under deletion the prediction degrades gradually, to $0.27$ at step 22 ($29\%$ of the pixels), while still covering the cat and spreading over the chair and the wall. It then fails within two steps: at step 24 ($32\%$) it has left the cat and covers almost only the background (mIoU $0.01$). Unlike SANSA's, the prediction does not vanish. Matcher merges the SAM masks that best match the support, so once nothing on the cat matches, it segments the regions that match the remaining evidence instead.

On the episode of \cref{fig:iauc_panet}, for the prototype head of \cref{sec:prototype_head}, the insertion curve grows almost linearly, from $0.32$ at step 13 to $0.58$ at the end, and keeps climbing after step 69, when the only pixels still hidden are the ones AffEx ranks lowest, outside the bicycles. The prototype head accounts for both: it averages the support features over the mask into a foreground and a background prototype, so each inserted pixel shifts an average instead of adding a match, and the pixels outside the object, ranked last by AffEx, still refine the background prototype. Deletion runs the other way: the mIoU stays between $0.54$ and $0.58$ up to step 5 and falls to zero at step 8 ($10.7\%$ of the pixels), once the bicycles in the supports are covered and the foreground prototype has almost nothing left to average.

Together, these examples show how IAUC and DAUC vary with episode difficulty and with the model's reliance on particular support pixels.

\clearpage

\end{document}